\documentclass[10pt,twocolumn,letterpaper]{article}

\usepackage{iccv}

\usepackage{times}
\usepackage{epsfig}
\usepackage{graphicx}
\usepackage{amsmath}
\usepackage{amssymb}
\usepackage{booktabs}
\usepackage{pifont}
\usepackage{caption}
\usepackage{soul}
\usepackage{multirow}
\usepackage{enumitem}
\usepackage{makecell}

\usepackage{inconsolata,enumitem}
\usepackage{subcaption}
\usepackage{xspace}
\usepackage{xcolor}

\usepackage[pagebackref=true,breaklinks=true,letterpaper=true,colorlinks,bookmarks=false]{hyperref}

\iccvfinalcopy 
\def\iccvPaperID{5428}

\newcommand{\ronghang}[1]{}

\newcommand{\aman}[1]{}

\newcommand{\amanadd}[1]{#1}
\newcommand{\amanswap}[2]{#2}

\newcommand{\modelNameWithoutSpace}{UniT}
\newcommand{\modelName}{\modelNameWithoutSpace\xspace}
\newcounter{magicrownumbers}
\newcommand\rownumber{\stepcounter{magicrownumbers}\arabic{magicrownumbers}}
\definecolor{demphcolor}{RGB}{124,124,124}
\newcommand{\demph}[1]{\textcolor{demphcolor}{#1}}

\begin{document}

\title{UniT: Multimodal Multitask Learning with a Unified Transformer}

\author{Ronghang Hu $\qquad$ Amanpreet Singh \\
Facebook AI Research (FAIR) \\
{\tt\small \{ronghanghu,asg\}@fb.com}
}

\twocolumn[{\renewcommand\twocolumn[1][]{#1}\maketitle
\centering
\small
\vspace{-2.5em}
\includegraphics[width=0.95\textwidth]{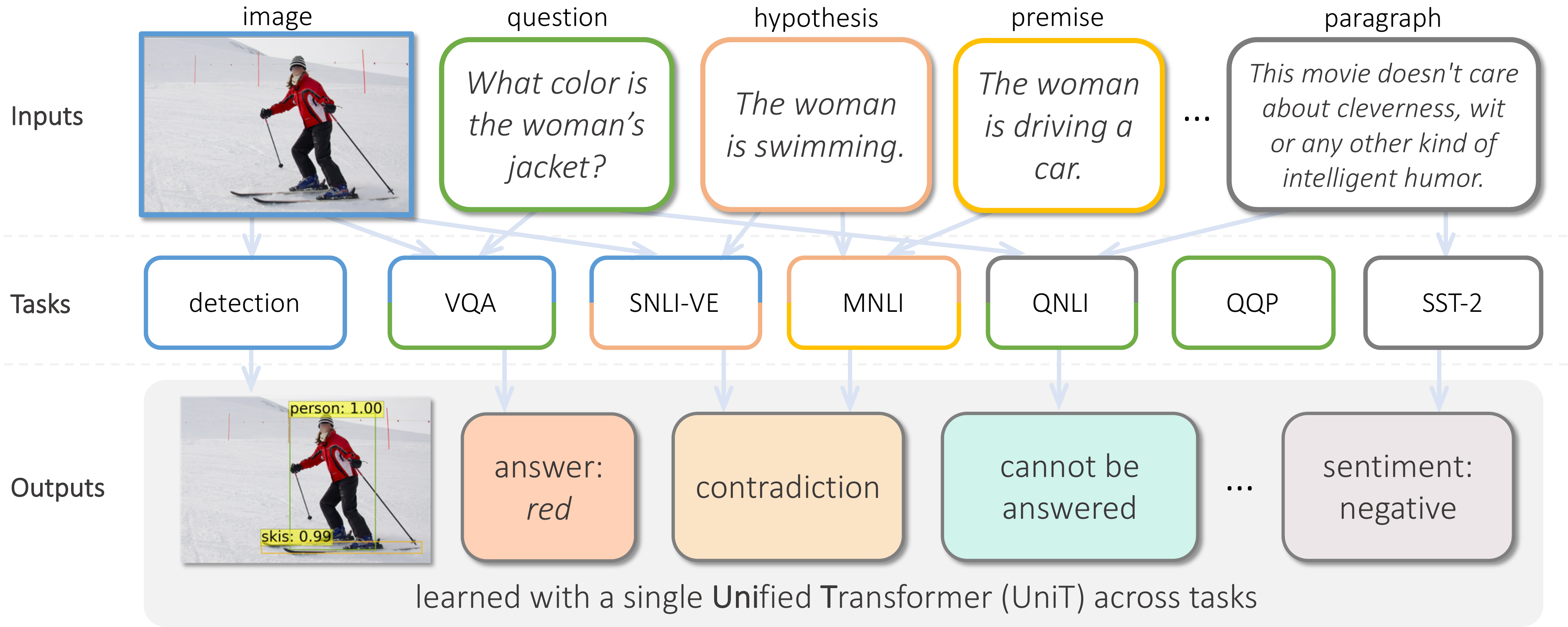} \\
\captionof{figure}{In this work, we propose \modelName, which jointly learns multiple tasks across different domains with a \textbf{Uni}fied \textbf{T}ransformer. Our \modelName model simultaneously handles 7 tasks on 8 datasets ranging from object detection to vision-and-language reasoning and natural language understanding, while achieving strong performance on each task with a compact set of model parameters.}
\label{fig:vis_popup}
\vspace{1em}
}]

\begin{abstract}
\vspace{-1em}
We propose UniT, a Unified Transformer model to simultaneously learn the most prominent tasks across different domains, ranging from object detection to natural language understanding and multimodal reasoning. Based on the transformer encoder-decoder architecture, our UniT model encodes each input modality with an encoder and makes predictions on each task with a shared decoder over the encoded input representations, followed by task-specific output heads. The entire model is jointly trained end-to-end with losses from each task. Compared to previous efforts on multi-task learning with transformers, we share the same model parameters across all tasks instead of separately fine-tuning task-specific models and handle a much higher variety of tasks across different domains. In our experiments, we learn 7 tasks jointly over 8 datasets, achieving strong performance on each task with significantly fewer parameters. Our code is available in MMF at \url{https://mmf.sh}.
\end{abstract}

\section{Introduction}
\label{sec:intro}

First proposed in \cite{vaswani2017attention}, transformers have shown great success in a wide range of domains including but not limited to natural language, images, video, and audio. Previous works (\eg \cite{devlin2018bert,radford2018improving,radford2019better,brown2020language,yang2019xlnet,liu2019roberta,lan2019albert,raffel2019exploring,rosset2020turing}) demonstrate that transformers trained on large corpora learn strong representations for a wide range of downstream language tasks. In the visual domain, models based on transformers have achieved promising results on image classification, object detection, and panoptic segmentation (\eg \cite{parmar2018image,bello2019attention,hu2018gather,hu2019local,ramachandran2019stand,dosovitskiy2020image,wang2018non,carion2020end,zhu2020deformable,beal2020toward,touvron2020training}). Besides modeling a single modality, transformer models also exhibit strong performance in joint vision-and-language reasoning tasks such as visual question answering (\eg \cite{li2019visualbert,lu2019vilbert,lu202012,Tan2019LXMERTLC,chen2020uniter,li2020unicoder,su2019vl,zhou2020unified,hu2020iterative}).

However, despite the above achievements in the application of transformers to \textit{specific domains}, there has not been much prior effort to connect different tasks \textit{across domains} with transformers. After witnessing the success of transformers, various questions naturally arise: could a transformer model trained for natural language inference on textual input also perform object detection on images, or could an image classifier based on transformers also check textual entailment? Overall, is it possible to build a single model that \textit{simultaneously} handles tasks in a variety of domains as a step towards general intelligence? Prior work tries to tackle some of these questions but only in limited scope:

\begin{itemize}[nosep,itemsep=1pt,leftmargin=1.5em,labelwidth=*,align=left]
    \item applied only to tasks from a single domain or specific multimodal domains; ViT \cite{dosovitskiy2020image} and DETR \cite{carion2020end} focus on vision-only tasks, BERT \cite{devlin2018bert} and its derivative works \cite{liu2019roberta,yang2019xlnet,lan2019albert,raffel2019exploring} only handle language tasks, while VisualBERT, VILBERT \cite{lu2019vilbert,li2019visualbert} and other multimodal transformers work only on specific multimodal domain of vision and language.
    \item involve task-specific fine-tuning for each of the tasks, not leveraging any shared parameters across the tasks, usually ending up with $\mathrm{N}$ times the parameters for $\mathrm{N}$ tasks, \eg one has to separately fine-tune a model for each of the tasks with BERT.
    \item perform multi-tasking upon related or similar tasks only from a single domain, sometimes with hard-coded training strategies; for example, T5 \cite{raffel2019exploring} works only on tasks in the language domain, while VILBERT-MT \cite{lu202012} works only on related vision-and-language tasks.
\end{itemize}
In this work, we build a \textbf{Uni}fied \textbf{T}ransformer \textbf{(\modelNameWithoutSpace)} model that takes images and/or text as inputs and jointly train on multiple tasks ranging from visual perception and natural language understanding to joint vision-\textit{and}-language reasoning. \modelName consists of transformer encoders which encode each input modality as a sequence of hidden states (feature vectors), and a transformer decoder over the encoded input modalities, followed by task-specific output heads applied on the decoder hidden states to make the final predictions for each of the tasks. Compared to previous work on multi-task learning with transformers (\eg~\cite{lu202012}), we train \modelName and achieve comparable performance to well-established prior work on a much larger variety of tasks; not only joint vision-and-language tasks such as visual question answering, but also vision-only as well as language-only tasks. We make the following contributions in this work:
\begin{itemize}[nosep,itemsep=1pt,leftmargin=1.5em,labelwidth=*,align=left]
    \item We propose \textbf{\modelName}, a \textbf{uni}fied \textbf{t}ransformer encoder-decoder architecture that handles multiple tasks and domains in a single model with fewer parameters, and a step towards general intelligence.
    \item We jointly learn the most prominent tasks in the visual and textual domains and their intersections, namely object detection, visual question answering (VQA), visual entailment, and natural language understanding tasks in the GLUE benckmark \cite{wang2019glue}, including QNLI~\cite{rajpurkar2016squad}, MNLI~\cite{williams2018broad}, QQP~\cite{qqp}, and SST-2~\cite{socher2013recursive}. We show that these diverse tasks can be learned simultaneously and converge properly under our training scheme.
    \item Through analyses across a variety of tasks, we show that multimodal tasks such as VQA and visual entailment benefit from multi-task training with uni-modal tasks.
\end{itemize}

\section{Related work}
\label{sec:related_work}

\paragraph{Transformers on language, vision, and multimodal tasks.} Transformers were first applied to the language domain for sequence-to-sequence modeling \cite{vaswani2017attention}. BERT \cite{devlin2018bert}, GPT \cite{radford2018improving,radford2019better,brown2020language}, XLNet \cite{yang2019xlnet}, RoBERTa \cite{liu2019roberta}, ALBERT~\cite{lan2019albert}, T5 \cite{raffel2019exploring}, T-NLG \cite{rosset2020turing} and other recent works show that transformers pretrained on large corpora learn language representations that can be transferred to a number of downstream tasks through fine-tuning.

In the visual domain, Image Transformer \cite{parmar2018image}, Image GPT \cite{chen2020generative}, DETR \cite{carion2020end}, ViT \cite{dosovitskiy2020image} and other recent works apply transformer models for several vision tasks. In addition, the multi-head self-attention mechanism from transformers also benefits a wide range of vision applications (\eg \cite{wang2018non,ramachandran2019stand,cordonnier2019relationship,zhao2020exploring,zhao2020point}). For joint vision-and-language reasoning tasks such as visual question answering, transformer models have been extended to take both the image and the text modalities as inputs (\eg VisualBERT \cite{li2019visualbert}, VILBERT \cite{lu2019vilbert,lu202012}, LXMERT \cite{Tan2019LXMERTLC}, and UNITER \cite{chen2020uniter}).

Most of these previous applications and extensions of transformers train (or fine-tune) a specific model for each of the tasks of interest. In BERT \cite{devlin2018bert}, a pretrained transformer model is fine-tuned separately on multiple downstream language tasks. In T5 \cite{raffel2019exploring}, a text-to-text transformer is jointly pretrained on different language tasks. However, despite learning generic representations through multi-task pretraining, T5 still fine-tunes a different set of parameters for each downstream task. On the contrary, we simultaneously learn multiple tasks within a single transformer. 

\begin{figure*}[t]
\vspace{-1.5em}
\centering
\includegraphics[width=0.75\linewidth]{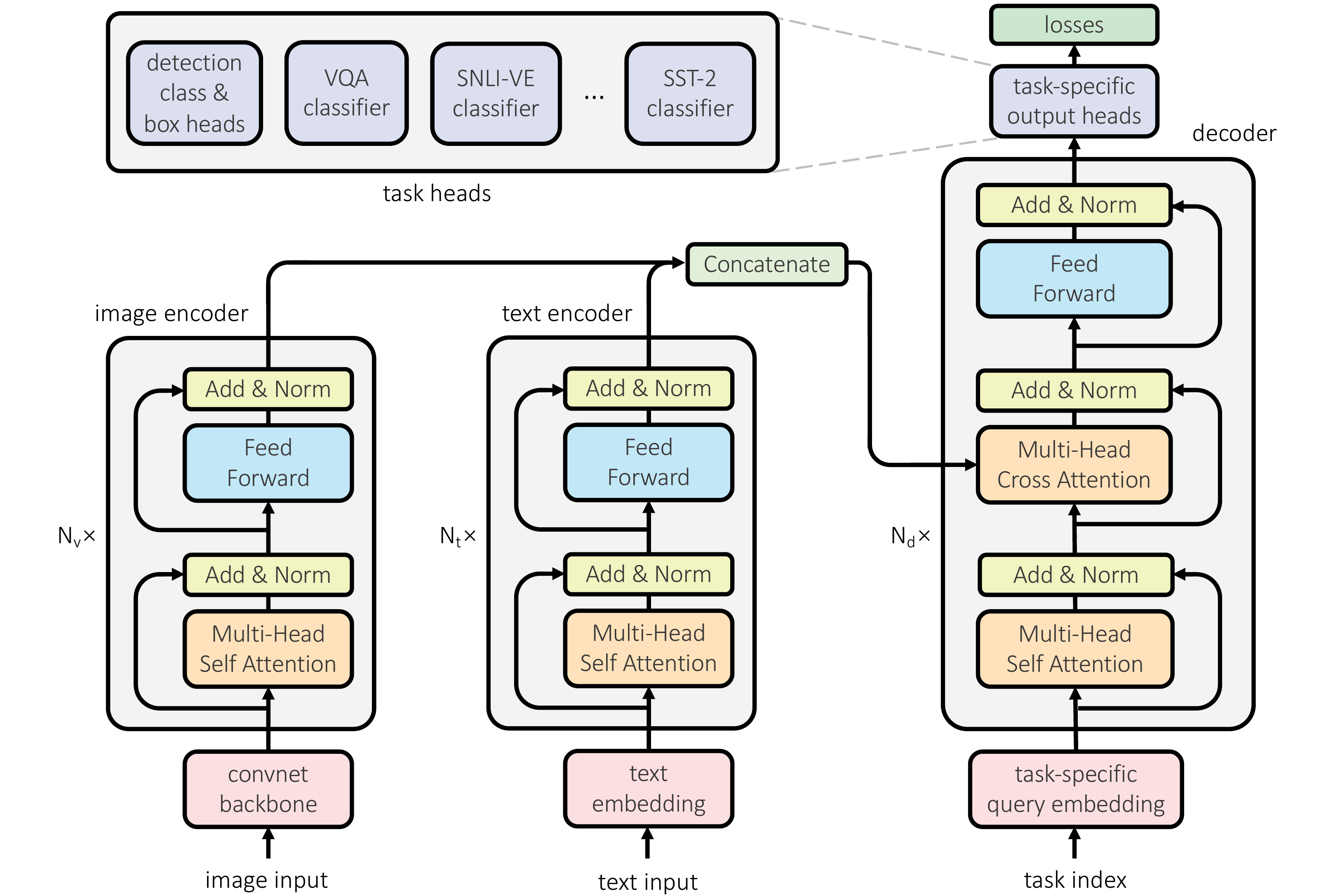}
\vspace{-0.5em}
\caption{An overview of our \modelName model, which jointly handles a wide range of tasks in different domains with a unified transformer encoder-decoder architecture. Our model uses an image encoder to encode the visual inputs (Sec.~\ref{sec:method_enc_image}), a text encoder to encode the language inputs (Sec.~\ref{sec:method_enc_text}), and a joint decoder with per-task query embedding (Sec.~\ref{sec:method_dec}) followed by task-specific heads (Sec.~\ref{subsec:task_specific_heads}) to make the final outputs for each task.}
\label{fig:method}
\vspace{-1.5em}
\end{figure*}

\vspace{-1em}
\paragraph{Multi-task learning with transformers.} There has been a long history of work on multi-task learning \cite{caruana1997multitask,crawshaw2020multi} in vision (\eg \cite{he2017mask,zamir2018taskonomy,strezoski2019many,standley2020tasks,zamir2020robust}), language (\eg \cite{sogaard2016deep,hashimoto2016joint,liu2017adversarial,sanh2019hierarchical,clark2019bam}), or multimodal areas (\eg \cite{kaiser2017one,kiela2018learning,pramanik2019omninet,chaplot2019embodied,lu202012}). Most previous efforts on multi-task learning focus on specific domains or modalities, often with model architectures tailored to the domain. However, there are also notable prior works on multi-task learning across domains with a single generic model. In \cite{kaiser2017one}, it is shown that an encoder-decoder architecture based on transformer's multi-head attention mechanism can be applied to different input and output domains such as image classification, machine translation, and image captioning. The decoders in \cite{kaiser2017one} are specifically designed for each output task, while our model involves fewer task-specific details as we apply the same decoder architecture on all tasks. In MT-DNN \cite{liu2019multi}, a multi-task language understanding model is built by sharing lower layers in a transformer while making the top layer task-specific. 
In VILBERT-MT \cite{lu202012}, 12 vision-and-language tasks were jointly learned with a multi-task transformer model based on VILBERT \cite{lu2019vilbert}. Compared to \cite{liu2019multi} and \cite{lu202012},  we expand beyond fixed input modalities and jointly handle different single-modal (vision-only and language-only) and multimodal tasks with a unified transformer model. In addition, our model allows end-to-end training directly over image pixels, instead of relying on pretrained detectors in \cite{lu202012}.

\vspace{-1em}
\paragraph{Contrast to multimodal pretraining.} Prior works such as VirTex \cite{desai2020virtex},  Voken \cite{tan2020vokenization} and VisualBERT \cite{li2019visualbert} show that pretraining on multimodal data such as image captions helps downstream vision, language, or multimodal tasks, which is often accomplished by building specialized models through fine-tuning on each downstream task. Unlike these approaches, we handle all tasks in a shared model, where the general knowledge across domains is not lost due to fine-tuning on specific downstream tasks. We believe the ability to jointly solve different tasks across domains is a critical step towards general intelligence.

\section{\modelName: Unified Transformer across domains}
\label{sec:method}

In this work, we jointly learn multiple tasks across different modalities with a unified single model. Our model, \modelName, is built upon the transformer encoder-decoder architecture \cite{vaswani2017attention,carion2020end}, consisting of separate encoders for each input modality type followed by a decoder (per-task or shared) with simple task-specific heads. Figure~\ref{fig:method} shows an overview of \modelName.

We consider two input modalities: images and text. For our transformer-based encoder on image inputs, inspired by \cite{carion2020end}, 
we first apply a convolutional neural network backbone to extract a visual feature map, which is further encoded by a transformer encoder into a list of hidden states to incorporate global contextual information. For language inputs, we use BERT \cite{devlin2018bert}\amanadd{, specifically the 12-layer uncased version,} to encode the input words (\eg questions) into a sequence of \amanswap{BERT hidden states}{hidden states from BERT's last layer}. After encoding input modalities into hidden state sequences, we apply the transformer decoder on either a single encoded modality or the concatenated sequence of both encoded modalities, depending on whether the task is uni-modal (\ie vision-only or language-only) or multimodal. We explore either having separate (\ie task-specific) \amanswap{decoders for each task, or sharing the same decoder for all tasks. 
On top of the decoder outputs, we apply simple task-specific output heads, such as a classifier, to obtain the final outputs for each task.}{or shared decoders among all tasks. Finally, the representation from the transformer decoder is passed to a task-specific head such as a simple two-layer classifier, which outputs the final predictions. Given the simplicity of \modelName, it can be extended easily to more modalities and inputs.}

We empirically show that our model can jointly learn 7 different tasks on 8 datasets. The following sections further describe the details of each component in \modelName.

\subsection{Image encoder}
\label{sec:method_enc_image}

The vision-only tasks (such as object detection) and vision-and-language tasks (such as visual question answering and visual entailment) require perceiving and understanding an image $I$ as input. In our model, we encode the input image $I$ with a convolutional neural network followed by a transformer encoder, into a list of encoded visual hidden states $\mathbf{h}^v = \{h^v_1, h^v_2, \cdots, h^v_L\}$.

Our image encoding process is inspired by DETR \cite{carion2020end}. First, a convolutional neural network backbone $B$ is applied on the input image to extract a visual feature map $\mathbf{x}^v$ of size $H_v \times W_v \times d_v^b$ as
\begin{equation}
    \mathbf{x}^v = B(I).
\end{equation}
In our implementation, the backbone network $B$ follows the structure of ResNet-50 \cite{he2016deep} with dilation \cite{yu2015multi} applied to its last C5 block, and is pretrained on object detection in \cite{carion2020end}.

We apply a visual transformer encoder $E_v$ with $N_v$ layers and hidden size $d_v^e$ on top of the feature map $\mathbf{x}^v$ to further encode it to visual hidden states $\mathbf{h}^v$ of size $L\times d_v^e$ (where $L = H_v \times W_v$ is the length of the encoded visual hidden states). In addition, given that different tasks (such as object detection and VQA) might require extracting different types of information, we also add a task embedding vector $w^{task}_v$ into the transformer encoder to allow it to extract task-specific information in its output as follows.
\begin{equation}
    \mathbf{h}^v = \left\{h^v_1, h^v_2, \cdots, h^v_L\right\} = E_v(P_{b\rightarrow e}(\mathbf{x}^v), w^{task}_v)
\end{equation}
$P_{b\rightarrow e}$ is a linear projection from visual feature dimension $d_v^b$ to encoder hidden size $d_v^e$. The structure of the visual transformer encoder $E_v$ follows DETR \cite{carion2020end}, where positional encoding is added to the feature map. The task token $w^{task}$ is a learned parameter of dimension $d_v^e$, which is concatenated to the beginning of the flattened visual feature list $P_{b\rightarrow e}(\mathbf{x}^v)$ and stripped from the output hidden states $\mathbf{h}^v$.

\subsection{Text encoder}
\label{sec:method_enc_text}

GLUE benchmark \cite{wang2019glue} tasks such as QNLI \cite{rajpurkar2016squad}, MNLI \cite{williams2018broad}, QQP \cite{qqp}, and SST-2 \cite{socher2013recursive} as well as the joint vision-and-language reasoning tasks such as VQA and visual entailment provide a textual input. 
We encode the textual input using BERT \cite{devlin2018bert} -- a transformer encoder model pretrained on large corpora with masked language modeling and next sentence prediction tasks.

Given the input text (\eg a sentence or a pair of sentences), we tokenize it in the same way as in BERT into a sequence of $S$ tokens $\{w_1, \cdots, w_S\}$, with $w_1=\text{[CLS]}$ (the special pooling token in BERT for classification). The token sequence is then used as input to a pretrained BERT model to extract a sequence of textual hidden states $\mathbf{h}^t$ of size $S \times d_t^e$, where $d_t^e$ is the BERT hidden size. Similar to the image encoder, in the text encoder, we also add a learned task embedding vector $w^{task}_t$ as part of the BERT input by prefixing it at the beginning of the embedded token sequence, and later stripping it from the output text hidden states as follows.
\begin{equation}
    \mathbf{h}^t = \left\{h^t_1, h^t_2, \cdots, h^t_S\right\} = \mathrm{BERT}(\{w_1, \cdots, w_S\}, w^{task}_t)
\end{equation}
However, we find that it works nearly equally well in practice to keep only the hidden vector corresponding to [CLS] in $\mathbf{h}^t$ as input to the decoder (which saves computation).

In our implementation, we use a pretrained BERT-base uncased model from the Huggingface's Transformers library~\cite{wolf2020huggingfaces}, which has $d_t^e=768$ and $N_t = 12$ layers.

\subsection{Domain-agnostic UniT decoder}
\label{sec:method_dec}

After encoding the input modalities, we apply on them a transformer decoder $D$ with hidden size $d^d_t$ and number of layers $N_d$ to output a sequence of decoded hidden states $\mathbf{h}^{dec}$ for predictions on each task. Unlike the image and text encoders with specific architectural designs for each modality, our decoder is built upon the same domain-agnostic transformer decoder architecture \cite{vaswani2017attention} across all tasks.

For vision-only tasks, we apply the decoder on the encoded image $\mathbf{h}^{enc} = \mathbf{h}^v$ described in Sec.~\ref{sec:method_enc_image}, for language-only tasks, we apply the decoder on the encoded text $\mathbf{h}^{enc} = \mathbf{h}^t$ in Sec.~\ref{sec:method_enc_text}, and finally for joint vision-and-language tasks, we concatenate the encoded inputs from both modalities into a single sequence $\mathbf{h}^{enc} = \mathrm{concat}(\mathbf{h}^v, \mathbf{h}^t)$ as the input to the decoder.

The transformer decoder $D$ takes the encoded input sequence $\mathbf{h}^{enc}$ and a task-specific query embedding sequence $\mathbf{q}^{task}$ of length $q$. It outputs a sequence of decoded hidden states $\mathbf{h}^{dec,l}$ for each of the $l$-th transformer decoder layer, which has the same length $q$ as the query embedding $\mathbf{q}^{task}$.
\begin{equation}
    \left\{\mathbf{h}^{dec,l}\right\} = D(\mathbf{h}^{enc}, \mathbf{q}^{task})
\end{equation}
Our decoder architecture mostly follows the transformer decoder implementation in DETR \cite{carion2020end}. In the $l$-th decoder layer, self-attention is applied among the decoder hidden states $\mathbf{h}^{dec,l}$ at different positions and cross-attention is applied to the encoded input modalities $\mathbf{h}^{enc}$.

In our experiments, we use either (i) a single shared decoder $D^{shared}$ for all tasks or (ii) a separate decoder $D^{sep}_{t}$ for each specific task $t$.

\subsection{Task-specific output heads}
\label{subsec:task_specific_heads}

A task-specific prediction head is applied over the decoder hidden states $\left\{\mathbf{h}^{dec,l}\right\}$ for each task $t$. For object detection, we use a class head to produce a classification output (including ``background'') and a box head to produce a bounding box output for each of the positions in $\{1,\dots,q\}$ in the decoder hidden states. The class head and the box head follow the implementation in DETR~\cite{carion2020end}. For datasets with attribute labels on each box (the Visual Genome dataset \cite{krishna2017visual} in our experiments), we also add an attribute classification head following the implementation of BUTD \cite{anderson2018bottom}. Each position in the decoder hidden states either produces an object class or background.

The outputs from the class and box heads are post-processed into object bounding boxes. Similar to \cite{carion2020end}, we apply these heads to all layers $l$ in the decoder hidden states $\mathbf{h}^{dec,l}$ during training as
\begin{eqnarray}
    \mathbf{c}^l &=& \mathrm{class\_head}(\mathbf{h}^{dec,l}) \\
    \mathbf{b}^l &=& \mathrm{box\_head}(\mathbf{h}^{dec,l}) \\
    \mathbf{a}^l &=& \mathrm{attr\_head}(\mathbf{h}^{dec,l}, \mathbf{c}^l)
\end{eqnarray}
where $\mathbf{c}^l$, $\mathbf{b}^l$, and $\mathbf{a}^l$ are class, box and attribute output sequences, all having the same length $q$ as the query embedding $\mathbf{q}^{task}$ for detection.

At test time, we only take the prediction from the top decoder layer, $\mathbf{h}^{dec,N_d}$. Since different detection datasets often have different numbers of classes, when training on multiple detection datasets, each dataset is given its own class, box, and attribute heads. We apply the same detection losses on the outputs $\mathbf{c}^l$ and $\mathbf{b}^l$ as in DETR~\cite{carion2020end}, and the same attribute losses on $\mathbf{a}^l$ as in BUTD \cite{anderson2018bottom}.

All other tasks that we address in this work, including visual question answering, visual entailment, and natural language understanding (QNLI, QQP, MNLI, and SST-2) can be cast as a classification task among $c_t$ classes for task $t$. We apply a task-specific classifier on the first output position hidden state $\mathbf{h}^{dec,N_d}_1$ from the top decoder layer to output a classification prediction $\mathbf{p}$ of size $c_t$ for the task $t$.

To predict the output classes, we use a two-layer MLP classifier with $\mathrm{GeLU}$ activation \cite{hendrycks2016bridging} (followed by dropout) and hidden dimension equal to decoder hidden size. We apply the cross-entropy classification loss on the predictions $\mathbf{p}$ with ground-truth targets $\mathbf{t}$ to train the model as follows.
\begin{eqnarray}
    \mathbf{p} &=& \mathrm{W}_1 \cdot \mathrm{GeLU}(\mathrm{W}_2 \cdot \mathbf{h}^{dec,N_d}_1 + \mathrm{b}_2) + \mathrm{b}_1 \\
    \mathrm{loss} &=& \mathrm{CrossEntropyLoss}(\mathbf{p}, \mathbf{t})
\end{eqnarray}

\subsection{Training}

We jointly train \modelName on multiple tasks. At each iteration during training, we randomly select a task and a dataset to fill a batch of samples. We manually specify a sampling probability for each task based on the dataset size and empirical evidence. In our implementation, we train with a batch size of 64 on 64 Nvidia Volta V100-SXM2-32GB GPUs (batch size 1 per GPU) in a distributed fashion, using 
PyTorch \cite{paszke2019pytorch}.

We use the weighted Adam optimizer \cite{kingma2014adam,loshchilov2019decoupled} with a learning rate of 5e-5 and the warm-up cosine learning rate schedule \cite{loshchilov2016sgdr} (using 2000 warm-up iterations). The optimizer updates the model parameters based on gradients from the task losses.

We apply the scale and crop augmentation following DETR \cite{carion2020end} on image inputs during training for object detection. In a detection training batch, an input image is randomly resized such that its shortest side is between 480 and 800 pixels, and then a crop with random width and height between 384 and 600 pixels is taken from the resized image. However, we do not apply scale and crop augmentation on vision-and-language tasks such as VQA, as these tasks often require the entire image for global reasoning (\eg answering ``how many people are there in the image'' requires counting every person in the entire image). At test time for object detection and at both training and test time for vision-and-language tasks, an input image is resized to have a deterministic shortest side of 800 pixels.

\section{Experiments}

To provide a thorough analysis of UniT and also provide a comparison with well-established prior work, we experiment with jointly learning prominent tasks from different domains, including object detection as a vision-only task, language understanding tasks from GLUE benchmark as language-only tasks, and also joint vision-and-language reasoning tasks. For object detection, we use the COCO dataset~\cite{lin2014microsoft} as a benchmark and also experiment with the Visual Genome (VG) dataset \cite{krishna2017visual}, which contains object classes as well as their attributes. For language understanding, we experiment with four tasks from the GLUE benchmark \cite{wang2019glue}: QNLI \cite{rajpurkar2016squad}, QQP \cite{qqp}, MNLI-mismatched~\cite{williams2018broad}, and SST-2 \cite{socher2013recursive}. For joint vision-and-language reasoning, we use the VQAv2 dataset \cite{goyal2017making} (with questions from Visual Genome \cite{krishna2017visual} as additional training data) and also experiment with the SNLI-VE dataset~\cite{xie2019visual}, which requires classifying an image and sentence pair into whether the sentence entails, contradicts or is neutral with respect to the image. These datasets are used for pure research purposes only.

We experiment with two settings. First, we jointly train our model on object detection and VQA tasks in Sec.~\ref{sec:exp_det_vqa}. Then, we further include language understanding tasks and SNLI-VE as an additional joint vision-and-language reasoning task in Sec.~\ref{sec:exp_all_tasks}.

\subsection{Multitask learning on detection and VQA}
\label{sec:exp_det_vqa}

We first experiment with training on object detection as a vision-only task and VQA as a multimodal task that requires jointly modeling the image and the text modalities.

\textbf{Removing overlap.} For object detection, we use the COCO detection dataset (COCO det.) \cite{lin2014microsoft} and the object annotations in the Visual Genome dataset (VG det.) \cite{krishna2017visual}. For the VQA task, we use the VQAv2 dataset \cite{goyal2017making}. We split these datasets according to COCO train2017 and val2017 splits: for COCO detection, we use its train2017 split for training and val2017 split for evaluation; for other datasets (Visual Genome detection and VQAv2), we train on those images not overlapping with COCO val2017 and evaluate on those images in COCO val2017. We also use those questions from the Visual Genome VQA dataset (on images not overlapping with COCO val2017) as additional training data, added to the training split of VQAv2.

\begin{table}[t]
\vspace{-1.5em}
\footnotesize
\setcounter{magicrownumbers}{0}
\setlength{\tabcolsep}{7pt}
\begin{center}
\begin{tabular}{@{}llrrr@{}}
\toprule
\multirow{2}[3]{*}{\bf~\#} & \multirow{2}[3]{*}{\bf~decoder setup} & \bf~COCO det. & \bf~VG det. & \bf~VQAv2 \\ 
 & & \bf~mAP & \bf~mAP & \bf~accuracy \\
\midrule
\rownumber & single-task training & 40.6 / ~~--~~~ & 3.87 & 66.38 / ~~~--~~~~ \\
\midrule
\rownumber & separate & \textbf{40.8} / ~~--~~~ & 3.91 & \textbf{68.84} / ~~~--~~~~ \\
\rownumber & shared & 37.2 / ~~--~~~ & 4.05 & 68.79 / ~~~--~~~~ \\
\rownumber & shared (COCO init.) & \textbf{40.8} / 41.1 & \textbf{4.53} & 67.30 / 67.47 \\
\bottomrule
\end{tabular}
\end{center}
\vspace{-2em}
\caption{Performance of \modelName on multi-task training over object detection and VQA. Our final model with a single shared decoder outperforms the separately trained single-task models on all the three datasets (line 4 vs line 1). On the COCO detection and VQAv2 datasets, we also evaluate on the test-dev splits for our final model.}
\label{tab:det_vqa_results}
\vspace{-1.5em}
\end{table}

\textbf{Training.} We train and evaluate our model under different combinations of tasks and datasets: COCO detection and VQAv2, Visual Genome (VG) detection and VQAv2, and all three datasets together. We also train it on a single dataset as a comparison.

We experiment with two settings in our transformer decoder: 1) separate decoders on different tasks (without sharing decoder parameters) and 2) a single shared decoder for all tasks. Following previous work in these two domains, we evaluate the detection performance with mean average precision (mAP) and the VQA task with VQA accuracy.\footnote{\href{https://visualqa.org/evaluation.html}{https://visualqa.org/evaluation.html}} During joint training, we sample all datasets with equal probability. We train for a total of 150k, 300k, and 450k iterations for experiments on one, two, and three datasets, respectively.\footnote{When training on multiple datasets jointly with shared decoders, we empirically find that skipping optimizer updates (including momentum accumulation) on unused parameters with zero gradients (\eg VQA classifier weights in a detection iteration) works better than updating all parameters. The latter often causes divergence, possibly because accumulating zero gradients leads to unstable momentum.}

\textbf{Results.} Table~\ref{tab:det_vqa_results} shows the performance of our model jointly trained on the three datasets with separate (line 2) or shared decoders (line 3), and also the single-task performance of our model trained separately on each dataset (line 1). With separate decoders, our model trained jointly on the three datasets outperforms its counterparts with single-task training on all three datasets. However, comparing line 3 with 1, we observe that while the joint model trained with shared decoders achieves better performance on VQA and VG detection, it underperforms the single-task models on COCO detection by a noticeable margin.

The object detection task requires structural outputs (bounding boxes with class labels, as opposed to a classification output in VQA), and the decoder needs to properly model the relations between different objects (such as their overlap to learn non-maximum suppression). Hence, object detection may require a longer training schedule, especially in the case of a single shared decoder, where the decoder needs to learn the complex behavior that models both the object relation in detection and the multimodal fusion and reasoning in VQA. To provide more training iterations on the detection task in the shared decoder setting, we experiment with initializing from a model trained on COCO detection alone (\textbf{COCO init.}) to continue training on the joint tasks. In this case, the image encoder (including the convolutional network backbone and the transformer encoder in it) and the detection heads are initialized from the single-task COCO detection model in Table~\ref{tab:det_vqa_results} line 1.
 
This variant of the joint model (in Table~\ref{tab:det_vqa_results} line 4) with shared decoders outperforms single-task models (line 1) on all three datasets. Also, comparing with line 3, it can be seen that the detection performance is notably better.\footnote{We find that the key to this improvement is to have sufficient training on the detection task, and an equivalent effect to COCO initialization can be achieved using $2\times$ total number iterations in joint training.}

We further evaluate with training on one dataset from each task (using either COCO or Visual Genome as the detection dataset). The results are shown in Table~\ref{tab:det_vqa_ablations}, where i) joint training on two detection datasets usually benefits both datasets (line 4 vs line 2 or 3) and ii) training on VG detection \& VQAv2 gives better VQA accuracy than training on COCO detection \& VQAv2 (line 3 vs 2), which is likely due to the fact that the Visual Genome dataset contains a more diverse set of object annotations (attributes) and better coverage of visual concepts for visual question answering.

\begin{table}[t]
\vspace{-1.5em}
\setcounter{magicrownumbers}{0}
\footnotesize
\begin{center}
\begin{tabular}{@{}llrrr@{}}
\toprule
\multirow{2}[3]{*}{\bf~\#} & \multirow{2}[3]{*}{\bf~training data} & \bf~COCO det. & \bf~VG det. & \bf~VQAv2 \\ 
 & & \bf~mAP & \bf~mAP & \bf~accuracy \\
\midrule
\rownumber & single-task training & 40.6 & 3.87 & 66.38 \\
\midrule
\rownumber & COCO + VQAv2 & 40.2 & -- & 66.88 \\
\rownumber & VG + VQAv2 & -- & 3.83 & \textbf{68.49} \\
\rownumber & COCO + VG + VQAv2 & \textbf{40.8} & \textbf{4.53} & 67.30 \\
\bottomrule
\end{tabular}
\end{center}
\vspace{-2em}
\caption{Object detection and VQA with shared decoders (COCO init.) on different dataset combinations. The two detection datasets benefit each other through joint training (line 4 vs line 2 or 3). Also, compared to COCO detection, VG detection has a larger benefit to VQA (line 3 vs 2).}
\label{tab:det_vqa_ablations}
\vspace{-1.5em}
\end{table}

\begin{table*}[t]
\vspace{-1.5em}
\setcounter{magicrownumbers}{0}
\footnotesize
\setlength{\tabcolsep}{4.5pt}
\begin{center}
\begin{tabular}{@{}llrrrrrrrr@{}}
\toprule
 &  & \bf~COCO det. & \bf~VG det. & \bf~VQAv2 & \bf~SNLI-VE & \bf~QNLI & \bf~MNLI-mm & \bf~QQP & \bf~SST-2 \\ 
\bf~\# & \bf~decoder setup & \bf~mAP & \bf~mAP & \bf~accuracy & \bf~accuracy & \bf~accuracy & \bf~accuracy & \bf~accuracy & \bf~accuracy \\
\midrule
\rownumber & UniT -- single-task training & 40.6 & 3.87 & 66.38 / ~~~--~~~~ & 70.52 / ~~~--~~~~ & 91.62 / ~~--~~~ & 84.23 / ~~--~~~ & 91.18 / ~~--~~~ & 91.63 / ~~--~~~ \\
\midrule
\rownumber & UniT -- separate & 32.2 & 2.54 & 67.38 / ~~~--~~~~ & 74.31 / ~~~--~~~~ & 87.68 / ~~--~~~ & 81.76 / ~~--~~~ & 90.44 / ~~--~~~ & 89.40 / ~~--~~~ \\
\rownumber & UniT -- shared & 33.8 & 2.69 & 67.36 / ~~~--~~~~ & 74.14 / ~~~--~~~~ & 87.99 / ~~--~~~ & 81.40 / ~~--~~~ & 90.62 / ~~--~~~ & 89.40 / ~~--~~~ \\

\rownumber & UniT -- separate (COCO init.) & 38.9 & 3.22 & 67.58 / ~~~--~~~~ & 74.20 / ~~~--~~~~ & 87.99 / ~~--~~~ & 81.33 / ~~--~~~ & 90.61 / ~~--~~~ & 89.17 / ~~--~~~ \\
\rownumber & UniT -- shared (COCO init.) & 39.0 & 3.29 & 66.97 / 67.03 & 73.16 / 73.16 & 87.95 / 88.0 & 80.91 / 79.8 & 90.64 / 88.4 & 89.29 / 91.5 \\
\midrule
\rownumber & \demph{UniT -- per-task finetuning} & \demph{42.3} & \demph{4.68} & \demph{67.60} / ~~~--~~~~ & \demph{72.56} / ~~~--~~~~ & \demph{86.92} / ~~--~~~ & \demph{81.53} / ~~--~~~ & \demph{90.57} / ~~--~~~ & \demph{88.06} / ~~--~~~ \\
\midrule
\rownumber & DETR \cite{carion2020end} & 43.3 & 4.02 & --~~~~~~~~~~ & --~~~~~~~~ & --~~~~~~~~ & --~~~~~~~~ & --~~~~~~~~ & --~~~~~~~~ \\
\rownumber & VisualBERT \cite{li2019visualbert} & --~~ & --~~ & 67.36 / 67.37 & 75.69 / 75.09 & --~~~~~~~~ & --~~~~~~~~ & --~~~~~~~~ & --~~~~~~~~ \\
\rownumber & BERT \cite{devlin2018bert} (bert-base-uncased) & --~~ & --~~ & --~~~~~~~~~~ & --~~~~~~~~ & 91.25 / 90.4 & 83.90 / 83.4 & 90.54 / 88.9 & 92.43 / 93.7 \\
\bottomrule
\end{tabular}
\end{center}
\vspace{-2em}
\caption{\textbf{Performance of our \modelName model on 7 tasks across 8 datasets}, ranging from vision-only tasks (object detection on COCO and VG), vision-and-language reasoning tasks (visual question answering on VQAv2 and visual entailment on SNLI-VE), and language-only tasks from the GLUE benchmark (QNLI, MNLI, QQP, and SST-2). For the line 5, 8 and 9, we also show results on VQAv2 test-dev, SNLI-VE test, and from GLUE evaluation server. See Sec.~\ref{sec:exp_all_tasks} for details.}
\label{tab:all_results}
\vspace{-1.5em}
\end{table*}

\begin{table}[b]
\vspace{-1.5em}
\footnotesize
\setcounter{magicrownumbers}{0}
\setlength{\tabcolsep}{3pt}
\begin{center}
\renewcommand{\arraystretch}{2}
\begin{tabular}{@{}ll@{}rrr@{}}
\toprule
\bf \# & \bf Model configuration & \makecell[r]{\bf~COCO det. \\\bf~mAP} & \makecell[r]{\bf~SNLI-VE\\\bf~accuracy} & \makecell[r]{\bf~MNLI-mm\\\bf~accuracy} \\ 
\midrule
\rownumber & \modelName (default, $d^d_t$=768, $N_d$=6 ) & 38.79 & 69.27 & 81.41 \\
\midrule
\rownumber & decoder layer number, $N_d$=8 & 40.13 & 68.17 & 80.58 \\
\rownumber & \makecell[l]{decoder layer number, $N_d$=12 \\\aman{400k iterations}}   & 39.02 & 68.82 & 81.15 \\
\rownumber & decoder hidden size, $d^d_t$=256   & 36.32 & 69.68 & 81.09 \\
\rownumber & \makecell[l]{using all hidden states from \\ BERT instead of just [CLS]} & 38.24 & 69.76 & 81.31 \\
\rownumber & \makecell[l]{losses on all decoder layers \\ for SNLI-VE and MNLI-mm} & 39.46 & 69.06 & 81.67 \\
\rownumber & \makecell[l]{no task embedding tokens} & 38.61 & 70.22 & 81.45 \\
\rownumber & \makecell[l]{batch size~=~32} & 35.03 & 68.57 & 79.62 \\
\bottomrule
\end{tabular}
\end{center}
\vspace{-2em}
\caption{Ablation analyses of our \modelName model with different configurations on COCO detection, SNLI-VE, and MNLI.}
\label{tab:ablations}
\vspace{-0.5em}
\end{table}

\subsection{A Unified Transformer for multiple domains}
\label{sec:exp_all_tasks}

To further test the capabilities of \modelName, we extend the training to 8 datasets, adding 4 language-only tasks from the GLUE benchmark (QNLI, QQP, MNLI, and SST-2) and a new vision-and-language dataset SNLI-VE for visual entailment. We show that \modelName can jointly perform on all 7 tasks across 8 datasets competitively using 8$\times$ fewer parameters than task-specific fine-tuned similar models. Our final \modelName model in Table~\ref{tab:all_results} line 5 has 201M parameters.

\textbf{Training.} For COCO, Visual Genome, and VQAv2, we follow the splits created in Sec.~\ref{sec:exp_det_vqa}. For SNLI-VE and the GLUE tasks, we follow the official splits.\footnote{GLUE tasks were downloaded from \href{https://gluebenchmark.com/tasks}{https://gluebenchmark.com/tasks}}\footnote{SNLI-VE was acquired from \href{https://github.com/necla-ml/SNLI-VE}{https://github.com/necla-ml/SNLI-VE}} Similar to Sec.~\ref{sec:exp_det_vqa}, we experiment with three different settings: (i) single-task training where each model is trained separately on each task, (ii) multi-task training with separate decoders where the model has a specific decoder for each task but is jointly trained on all of the tasks, and (iii) multi-task training same as (ii) but with a shared decoder instead of separate ones. In (iii), the model still contains lightweight task-specific heads for each task to generate predictions as explained in Sec.~\ref{subsec:task_specific_heads}. Following Sec.~\ref{sec:exp_det_vqa}, we also train a variation of (ii) and (iii), where we initialize the image encoder and the decoder from a single task COCO-pretrained \modelName model (referred to as COCO init.). We train all models for 500k iterations and keep the rest of the hyper-parameters the same as in previous experiments in Sec.~\ref{sec:exp_det_vqa}.

\textbf{Results.} Table~\ref{tab:all_results} shows the performance of \modelName under different variants. Here, the \modelName models trained on each task separately (line 1) outperform all other variants (line 2 to 4) on all tasks except multimodal tasks VQAv2 and SNLI-VE. This is unsurprising as (i) the unimodal tasks have low cross-modality overlap, 
(ii) in joint training, each task is trained only for a proportion of the total training iterations, 
and (iii) the shared decoder (line 3 and 5) has 8$\times$ fewer parameters compared to the models in line 1. On the other hand, we see that vision-and-language tasks, namely VQAv2 and SNLI-VE, consistently benefit from multi-task training together with vision-only and language-only tasks across different settings, suggesting that learning better uni-modal representations also benefits multimodal reasoning.

\begin{figure*}[t]
\vspace{-1em}
\centering
\includegraphics[width=0.95\linewidth]{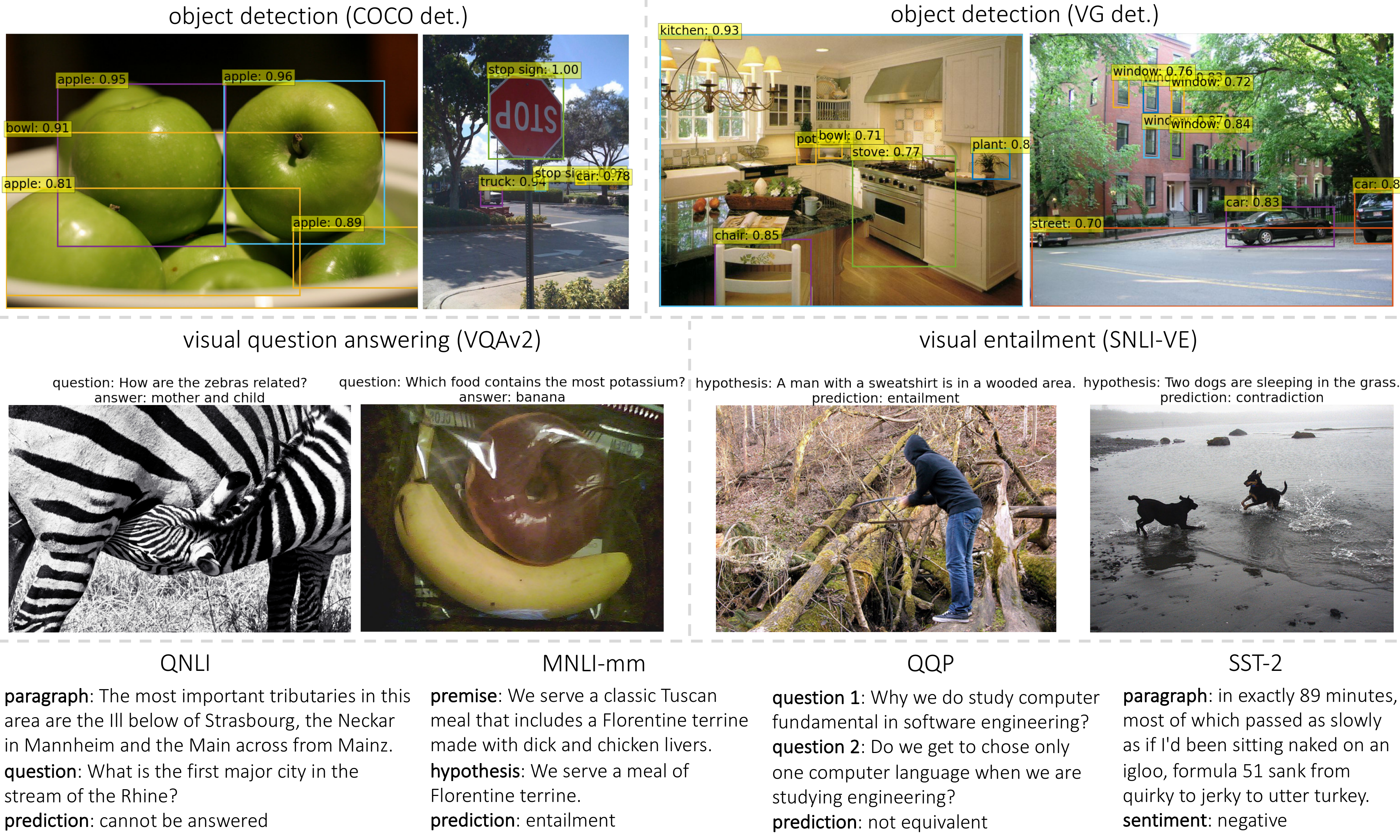}
\vspace{-0.8em}
\caption{Predictions of our model with a shared decoder (Table~\ref{tab:all_results} line 5) across 8 datasets. Our model jointly handles a large variety of tasks above through a unified transformer encoder-decoder architecture.}
\label{fig:visualization}
\end{figure*}

In addition, we further explore fine-tuning our shared model (line 5) on each task and find that while per-task fine-tuning brings a notable boost to object detection, it only has a moderate impact and sometimes even a small drop on other tasks as shown in line 6. Note that despite better mAP on detection, per-task fine-tuning leads to $8\times$ more parameters, longer training, and loss of generality, which we would like to avoid since our goal is to build a general model.

\textbf{Comparison to previous work.} We compare \modelName to well-established domain-specific methods based on transformers on each task. For object detection, we compare to DETR \cite{carion2020end} (line 7), a recent transformer-based detector from which our image encoder is inspired. For joint vision-and-language reasoning (visual question answering and visual entailment), we compare to VisualBERT \cite{li2019visualbert} (line 8), which extends BERT \cite{devlin2018bert} to also take detected objects as inputs.\footnote{We compare to the variant of VisualBERT without masked language modeling pretraining on vision-and-language datasets for fair comparison.} On natural language understanding tasks from the GLUE benchmark, we compare to BERT \cite{devlin2018bert} (line 9). From Table~\ref{tab:ablations}, it can be seen that our model achieves strong performance on each task with a single generic model. Although there is still a gap when comparing line 5 to line 7, 8, and 9, our model shows promising results approaching these domain-specific transformer-based models -- especially considering that these previous approaches have hyperparameters tailored to each domain, while our model adopts the same hyperparameters across all 8 datasets. It also simplifies the training process as our whole model is trained end-to-end in one step for all tasks, while BERT and VisualBERT need to be separately trained on each task and VisualBERT also requires first training an external Faster R-CNN object detector \cite{ren2015faster}. Figure~\ref{fig:visualization} shows the predictions of our model (in Table~\ref{tab:all_results} line 5) on each dataset.

\textbf{Ablations.} To better understand the effect of each hyper-parameter on multi-modal multi-task training with \modelName, we conduct a range of ablations shown in Table~\ref{tab:ablations}. We choose one dataset from each domain: COCO for vision-only, SNLI-VE for vision-and-language, and MNLI for language-only. MNLI-mismatched and SNLI-VE are related tasks involving natural language inference at the core. Please see supplemental for more ablation analyses.

\begin{itemize}[nosep,itemsep=1pt,leftmargin=1em,labelwidth=*,align=left]
    \item \textbf{Decoder layers and hidden size:} There is a drop in detection mAP with a smaller decoder hidden size (line 4), while it does not hurt SNLI-VE or MNLI-mm. This is likely because COCO is a larger dataset with 1.5 million object instances and benefits from larger models. The analyses on decoder layer number $N_d$ (line 2 and 3) confirms this intuition as $N_d = 8$ gives better detection mAP. Meanwhile, doubling the decoder layers to $N_d = 12$ does not help detection as much, likely due to overfitting with very large models. In addition, we find that too large decoder hidden size ($d^d_t=1536$) could lead to divergence in detection training.
    \item \textbf{All hidden states in language encoder}: Using all BERT outputs as inputs to the decoder (instead of just the [CLS] token as in Sec.~\ref{sec:method_enc_text}) has a relatively minor (and mixed) impact on the performance while increasing computation cost (line 5), suggesting that the pooled vector from BERT should be sufficient for most downstream tasks.
    \item \textbf{Losses on all decoder layers}: While losses on intermediate layer outputs benefit object detection (as shown in \cite{carion2020end}), it does not benefit SNLI-VE or MNLI (line 6), likely because these tasks only require outputting a single label, unlike dense detection outputs.
    \item \textbf{No task embedding tokens}: We find that removing the task embedding from the encoders (line 7) does not hurt the performance. We suspect it is because the image encoder can extract generic (instead of task-specific) visual representations applicable to both COCO and SNLI-VE, and likewise for the language encoder.
    \item \textbf{Batch size and learning rate:} We find that a smaller batch size (line 8) leads to lower performance. In addition, we also find that a larger learning rate (1e-4 as in DETR \cite{carion2020end} and MLM in BERT \cite{devlin2018bert}) often causes divergence in joint training, while our smaller 5e-5 learning rate provides stable training.
\end{itemize}

\section{Conclusion}

In this work, we show that the transformer framework can be applied over a variety of domains to jointly handle multiple tasks within a single unified encoder-decoder model. Our \modelName model simultaneously addresses 7 tasks across 8 datasets, learning them in a single training step and achieving strong performance on each task with a compact set of shared parameters. Through a domain-agnostic transformer architecture, our model makes a step towards building general-purpose intelligence agents capable of handling a wide range of applications in different domains, including visual perception, natural language understanding, and reasoning over multiple modalities.

\noindent\textbf{Acknowledgments.} We are grateful to Devi Parikh, Douwe Kiela, Marcus Rohrbach, Vedanuj Goswami, and other colleagues at FAIR for fruitful discussions and feedback.

{\small
\bibliographystyle{ieee_fullname}
\bibliography{unit}
}

\appendix

\counterwithin{figure}{section}
\counterwithin{table}{section}
\counterwithin{equation}{section}

\twocolumn[{
\begin{center}
\Large 
\textbf{UniT: Multimodal Multitask Learning with a Unified Transformer}\\(Supplementary Material)
\par
\end{center}
\vspace{2em}
}]

\section{Hyper-parameters and details of \modelName}

\begin{table}[b]
\vspace{-1em}
\small
\begin{center}
\begin{tabular}{@{}lr@{}}
\toprule
\textbf{Hyper-parameter} & \textbf{Value} \\
\midrule
image encoder hidden size & 256 \\
image encoder head number & 8 \\
image encoder intermediate size & 2048 \\
image encoder layer number & 6 \\
image encoder dropout & 0.1 \\
decoder hidden size & 768 \\
decoder head number & 8 \\
decoder intermediate size & 2048 \\
decoder layer number & 6 \\
decoder dropout & 0.1 \\
\midrule
batch size & 64 \\
learning rate & 5e-5 \\
learning schedule & \texttt{warmup\_cosine} \\
warmup iterations & 2000\\
Adam $\beta_1$ & 0.9 \\
Adam $\beta_2$ & 0.999 \\
\bottomrule
\end{tabular}
\end{center}
\vspace{-1.5em}
\caption{A list of hyper-parameters in \modelName.}
\label{tab:supp_hyperparameters}
\end{table}

\begin{table*}[t]
\setcounter{magicrownumbers}{0}
\footnotesize
\setlength{\tabcolsep}{9pt}
\begin{center}
\begin{tabular}{@{}llrrrrrrrr@{}}
\toprule
\bf~\# & \bf~Experimental setting & \bf~COCO det. & \bf~VG det. & \bf~VQAv2 & \bf~SNLI-VE & \bf~QNLI & \bf~MNLI-mm & \bf~QQP & \bf~SST-2 \\ 
\midrule
\rownumber & detection + VQA (Sec.~4.1) & 0.33 & 0.33 & 0.33 & -- & -- & -- & -- & -- \\
\rownumber & all 8 tasks (Sec.~4.2) & 0.20 & 0.07 & 0.26 & 0.12 & 0.10 & 0.10 & 0.10 & 0.05 \\
\rownumber & ablation study (Sec.~4.2) & 0.30 & -- & -- & 0.50 & -- & 0.20 & -- & -- \\
\bottomrule
\end{tabular}
\end{center}
\vspace{-1.5em}
\caption{Sampling probabilities of each dataset for joint training under different experimental settings.}
\label{tab:supp_sampling_ratio}
\vspace{-1em}
\end{table*}

We summarize the hyper-parameters in our \modelName model in Table~\ref{tab:supp_hyperparameters}. We also list the sampling probabilities of each dataset during joint training in Table~\ref{tab:supp_sampling_ratio} under different experimental settings.

\vspace{-1em}
\paragraph{Unused parameters in the optimizer.} Some parameters in our model (\eg the task-specific output heads) are only used on a subset of tasks and datasets. During development, we first tried updating all parameters in the model during training even if some parameters were not used in the forward pass of a batch and their gradients remained zero. However, we empirically found that this strategy sometimes caused the training to diverge. On the other hand, the alternative strategy of skipping optimizer updates (including momentum accumulation) on unused parameters in a batch with zero gradients provides more stable training -- however, in some cases, this alternative training strategy yields slightly lower scores (\eg $-0.2\%$ lower accuracy on VQAv2).

When jointly training on COCO detection, VG detection, and VQAv2 with a shared decoder (Sec. 4.1 in the main paper), divergence happens if we update unused parameters in the optimizer, where the VQA accuracy stays around 25\%. The divergence might be related to a high overall sampling probability on detection (0.667), such that the detection gradients dominate the model. We find that the alternative strategy (skipping unused parameters in optimizer) allows the model to converge properly in this case. Meanwhile, lowering sampling probabilities on detection datasets also avoids such divergence on VQA, but gives lower detection mAP than this alternative strategy.

\section{Multitask learning in UniT}

In this work, we propose UniT -- a multi-task joint model across several domains achieving comparable performance to per-task models with $8\times$ fewer parameters. As discussed in Sec. 2 in the main paper, our model is notably different from previous work in the pretrain-and-transfer paradigm -- UniT is a joint and shared model instead of separately fine-tuned ones.

While per-task fine-tuning could be useful for single-task performance (and its results show that UniT can achieve competitive single-task performance), it is not ideal towards this multi-task goal, as one needs to save 8 separately fine-tuned models to handle all 8 tasks, leading to $8\times$ total parameters compared to a single shared UniT model.

In Table 3 in the main paper, our multi-task model (line 5) achieves better performance on VQAv2 and SNLI-VE but does not outperform separately-trained single-task models on pure vision or pure language tasks in line 1. We note that while multi-task learning sometimes benefits individual tasks, there is not much prior evidence on vision-and-language tasks helping pure vision tasks in a joint model via multi-task learning (instead of pretraining). In particular, no prior work to the best of our knowledge shows VQA, as compared to captioning, helps object detection via multi-task learning. Rather, better VQA accuracy often comes at sacrificing detection performance as detectors used in VQA are heavily specialized, \eg the detector trained in BUTD \cite{anderson2018bottom} has relatively poor localization performance on COCO classes.\footnote{on COCO classes: $15.2$ mAP@IoU=0.5 and $5.0$ mAP@IoU=0.5:0.95} Meanwhile, we handle \textit{both} detection and VQA with strong and comparable performance to prior work. Similarly, on vision-and-language and pure language tasks, we find that VisualBERT \cite{li2019visualbert} has a noticeable drop on GLUE accuracy\footnote{drop on QNLI, MNLI, QQP, SST-2: $-2.76$, $-2.50$, $-0.70$, $-2.06$} 
over the original BERT, while our model solves vision-and-language tasks, GLUE as well as detection jointly with reasonable performance.

We emphasize that UniT handles all tasks in a shared model, where knowledge on object detection and language is not lost due to specializing to other tasks, in contrast to prior work on pretrain-and-transfer. We believe UniT's ability to jointly solve different tasks across domains is a critical step towards general intelligence.

Also in our experiments, we show that UniT can be applied over a diverse set of tasks through a shared model, even if some of them are usually considered unrelated (such as object detection in vision and sentiment analysis in language). This confirms that task compatibility is not a strict requirement for UniT to learn a joint shared model. On the other hand, we also find that some tasks are more compatible than others for joint training. There are both benefits from joint multi-task learning (because they can share supervision) and competitions between tasks (due to a finite model capacity). Given this intuition, we find that it is often helpful to include more relevant and compatible tasks based on prior knowledge (\eg VQA benefits from better object detection) or a systematic taskonomy evaluation.\footnote{such as \url{http://taskonomy.stanford.edu/}}

\section{Additional ablation results}

\begin{table}[b]
\footnotesize
\setcounter{magicrownumbers}{0}
\setlength{\tabcolsep}{2.5pt}
\begin{center}
\renewcommand{\arraystretch}{2}
\begin{tabular}{@{}ll@{}rrr@{}}
\toprule
\bf \# & \bf Model configuration & \makecell[r]{\bf~COCO det. \\\bf~mAP} & \makecell[r]{\bf~SNLI-VE\\\bf~accuracy} & \makecell[r]{\bf~MNLI-mm\\\bf~accuracy} \\ 
\midrule
\rownumber & \modelName (default, $d^d_t$=768, $N_d$=6 ) & 38.79 & 69.27 & 81.41 \\
\midrule
\rownumber & \makecell[l]{image encoder hidden size, \\$d^e_v$=768}  & 33.39 & 68.53 & 81.01 \\
\rownumber & \makecell[l]{initializing backbone from \\ImageNet instead of DETR} & 36.65 & 69.07 & 80.64 \\

\rownumber & \makecell[l]{number of queries=1 \\ for SNLI-VE and MNLI-mm} & 38.75 & 68.66 & 81.66 \\
\rownumber & \makecell[l]{number of queries=100 \\ for SNLI-VE and MNLI-mm} & 38.63 & 69.14 & 81.09 \\
\rownumber & \makecell[l]{learning rate=1e-4}  & \multicolumn{3}{r}{(training diverged in this setting)} \\
\rownumber & learning rate=1e-5   & 29.88 & 70.39 & 83.74 \\
\rownumber & train for 1M iterations   & 39.96 & 69.31 & 79.88 \\
\rownumber & \makecell[l]{init from COCO single-task} & 40.98 & 68.72 & 81.08 \\
\rownumber & \makecell[l]{init from COCO single-task \\  w/ frozen encoders} & 38.88 & 65.77 & 61.47 \\
\rownumber & \makecell[l]{similar to 10 but do not init. \\ detection class and box heads} & 37.18 & 65.01 & 59.87 \\
\rownumber & \makecell[l]{similar to 10 but only freeze \\ vision encoder} & 37.87 & 68.70 & 81.11 \\
\bottomrule
\end{tabular}
\end{center}
\caption{Additional ablation analyses of our \modelName model with different model configurations on COCO detection, SNLI-VE, and MNLI (under the same settings as in Sec. 4.2 in the main paper).}
\label{tab:supp_ablations}
\end{table}

In Table~\ref{tab:supp_ablations}, we show more ablation results of our \modelName model on the three datasets, COCO detection, SNLI-VE, and MNLI, under the same settings as in our ablation analyses in Sec. 4.2 and Table 4 in our main paper:

\begin{itemize}[nosep,itemsep=1pt,leftmargin=1em,labelwidth=*,align=left]
    \item \textbf{Image encoder hidden size:} Increasing the hidden size of the image encoder from 256 (default in DETR) to 768 (the BERT hidden size) leads to noticeably lower detection performance (line 2), which is possibly due to overfitting in the detection features.
    \item \textbf{Initializing convnet backbone from ImageNet}: Instead of initializing the convolutional network backbone in the image encoder from a detection-pretrained ResNet-50 in DETR \cite{carion2020end}, in this setting (line 3) the backbone is initialized from a ResNet-50 pretrained on ImageNet classification. It can be seen that the classification-pretrained backbone leads to lower COCO detection mAP. We suspect this is due to a relatively small number of training iterations on the COCO detection dataset -- here we are using a total of 500k iterations on three datasets, while DETR \cite{carion2020end} is trained for over 900k iterations (500 epochs) on the COCO dataset alone.
    \item \textbf{The number of queries in decoder:} In this setting, we vary the number of the query vectors in the decoder (\ie the length of the query embedding sequence $\mathbf{q}^{task}$ in Sec.~3.3) on SNLI-VE and MNLI (while keeping a fixed number of 100 queries on the COCO detection task). We found that using only 1 query in the decoder (line 4) results in slightly lower accuracy on SNLI-VE, which is likely due to that the decoder needs to fuse multiple modalities in this case for visual entailment reasoning and benefits from more input queries. However, increasing the query number to 100 (line 5) does not give higher accuracy on SNLI-VE than the default setting (25 queries).
    \item \textbf{Learning rate:} We found that the joint training performance is sensitive to the learning rate. In line 6, training diverges with a higher learning rate (1e-4) than the default value of 5e-5. On the other hand, with a lower learning rate (1e-5) in line 7, the COCO detection mAP is noticeably lower while the SNLI-VE and MNLI accuracies are higher. These results show that different tasks have different optimal learning rates, which adds to the difficulties of joint training. Our default setting (line 1) uses a 5e-5 learning rate as a balance across tasks. A possible future direction is to explore custom and adaptive learning rates on different components of the model.
    \item \textbf{More training iterations:} Using $2\times$ more training iterations (1M) yields higher COCO detection mAP but lower MNLI accuracy (line 8). We suspect it is because the detection task requires a longer training schedule to output a list of boxes and classes, while the MNLI dataset only requires a single classification prediction and too many iterations could cause overfitting.
    \item \textbf{Initialization from the COCO single-task model:} To provide more training iterations on the detection task, in line 9 we also experiment with initializing the multi-task model from the single-task model trained on the COCO detection dataset alone (\ie COCO init. as described in Sec. 4.1 in the main paper). As expected, initializing from a COCO-pretrained single-task model leads to a noticeably higher detection mAP (line 9 vs 1), but we also see a slight performance drop on the other two datasets.
    \item \textbf{Freezing the encoders in \modelName:} In multi-task training with \modelName, the image and text encoders are jointly trained with the rest of the model. However, one might wonder whether it is necessary or beneficial to train these modality-specific encoders jointly. Is it possible to learn the encoders once on individual uni-modal tasks and directly use them on other tasks without retraining? \\
    \phantom{x}~~In this setting, we experiment with pretrained and frozen encoders. In line 10, we initialize the image encoder from a single-task model pretrained on COCO detection (same as in line 9), initialize the text encoder from a pretrained BERT model (bert-base-uncased), and freeze both decoders during training. We also train another variant (line 11), which is similar to line 10 except that the detection class and box heads are randomly initialized. \\
    \phantom{x}~~It can be seen that these two variants have significantly lower performance on all three datasets. In line 12, we still freeze the image encoder but update the text encoder (BERT) during training. It leads to better accuracy on MNLI and SNLI-VE that involve language understanding, but still relatively low detection mAP on COCO. These results suggest that it is hard to build a single shared decoder upon the frozen representations of each modality and that the co-adaptation of the decoder \textit{and} the encoders is critical to multi-task training.
\end{itemize}

\section{Learning curves}

\begin{figure*}
\vspace{-1.5em}
    \centering

    \begin{subfigure}[b]{0.8\linewidth}
    \includegraphics[width=\linewidth]{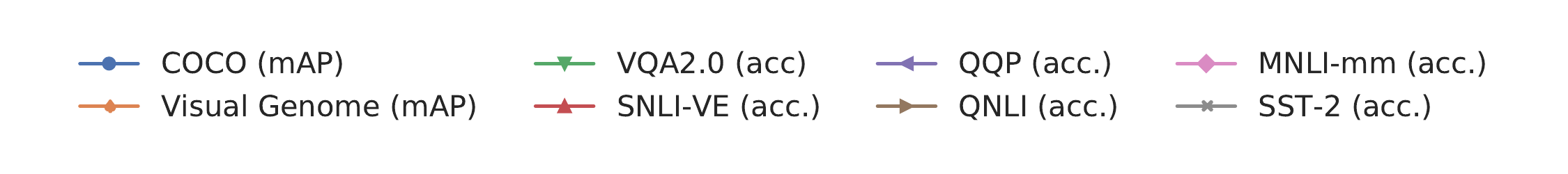}
    \end{subfigure}
        
    \begin{subfigure}[b]{0.3\linewidth}
    \includegraphics[width=\linewidth]{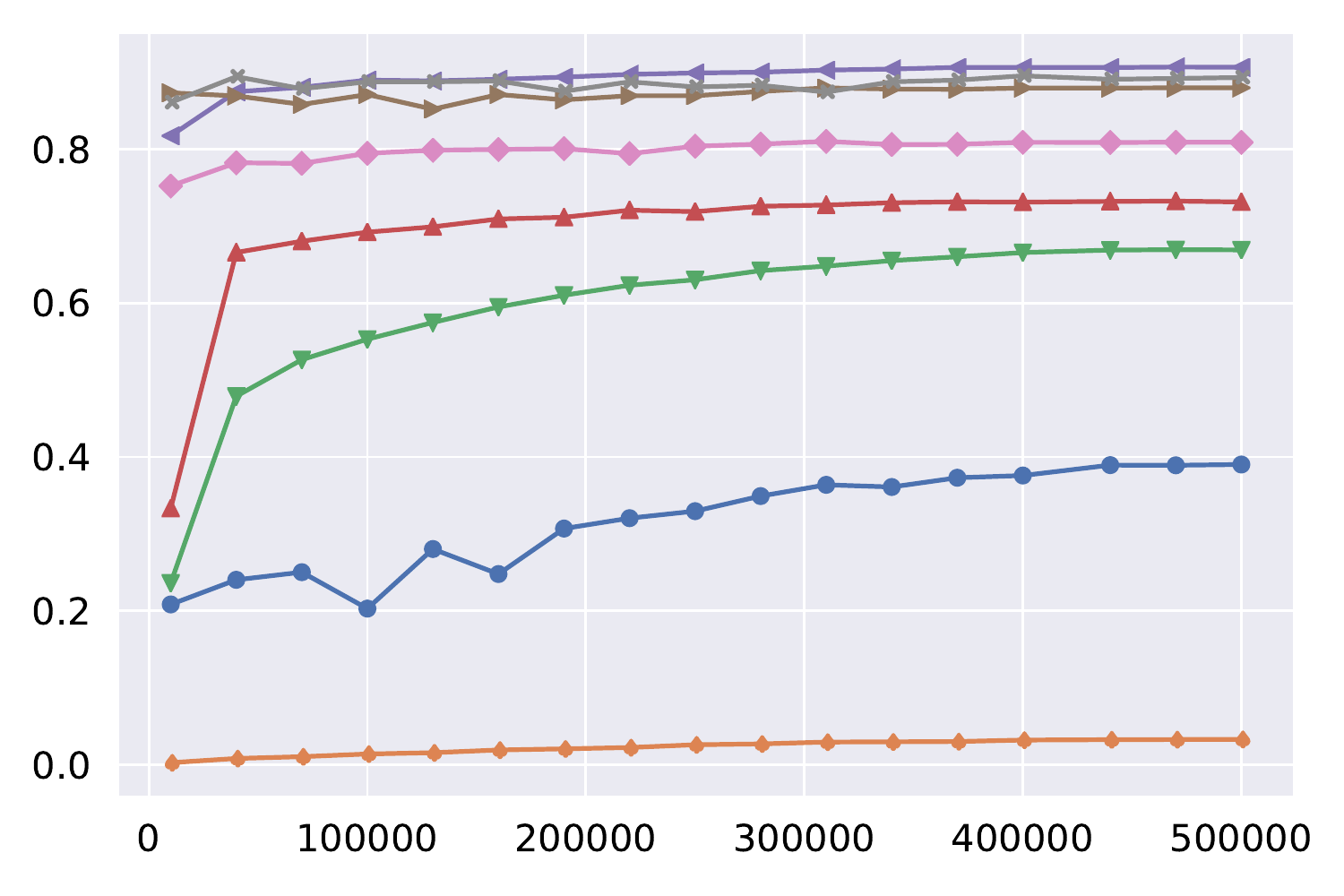}
    \caption{Shared decoders}
    \label{fig:shared_decoders_curve}
    \end{subfigure}
    \hspace{0.08\linewidth}
    \begin{subfigure}[b]{0.3\linewidth}
    \includegraphics[width=\linewidth]{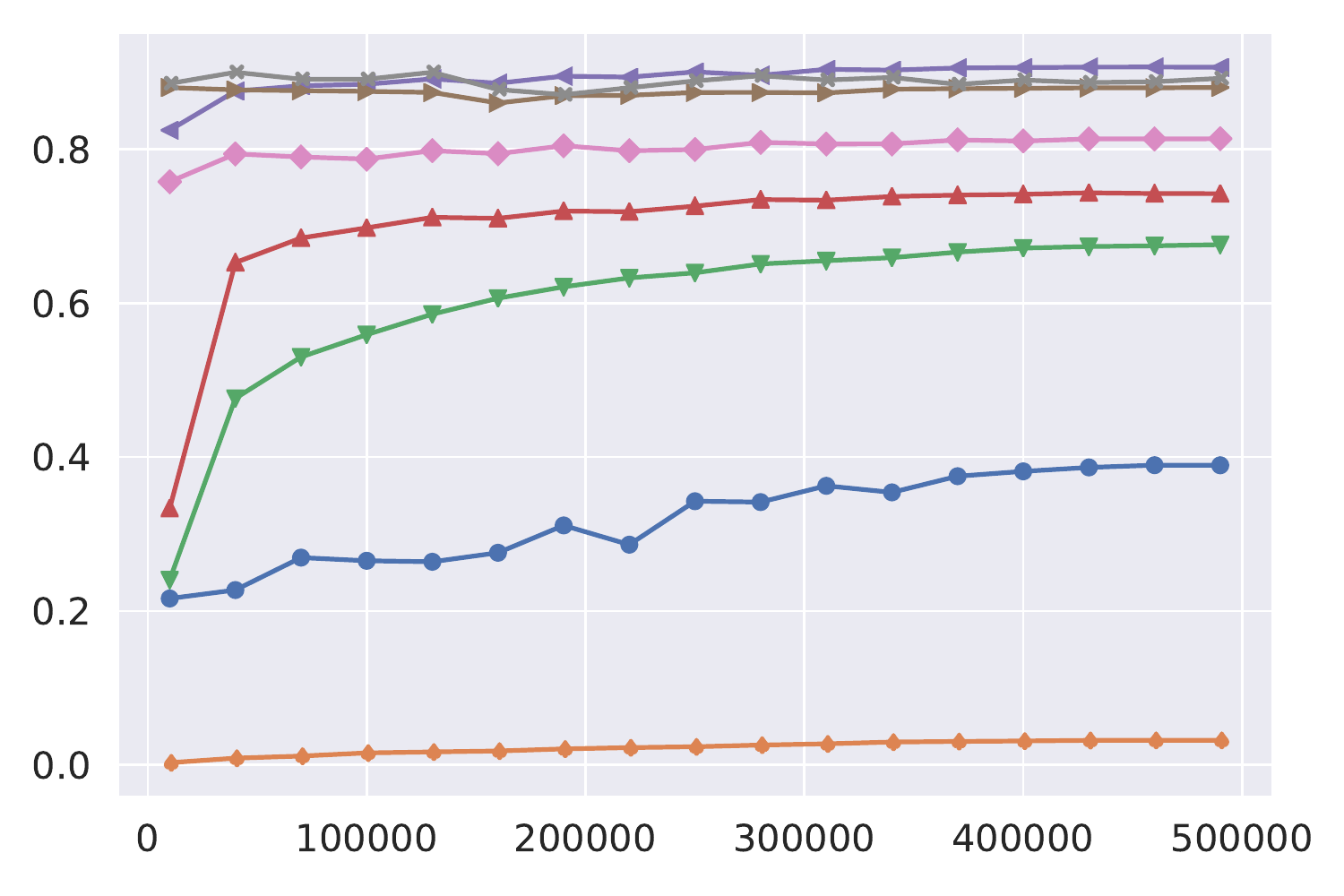}
    \caption{Separate decoders}
    \label{fig:separate_decoders_curve}
    \end{subfigure}

    \begin{subfigure}[b]{0.3\linewidth}
    \includegraphics[width=\linewidth]{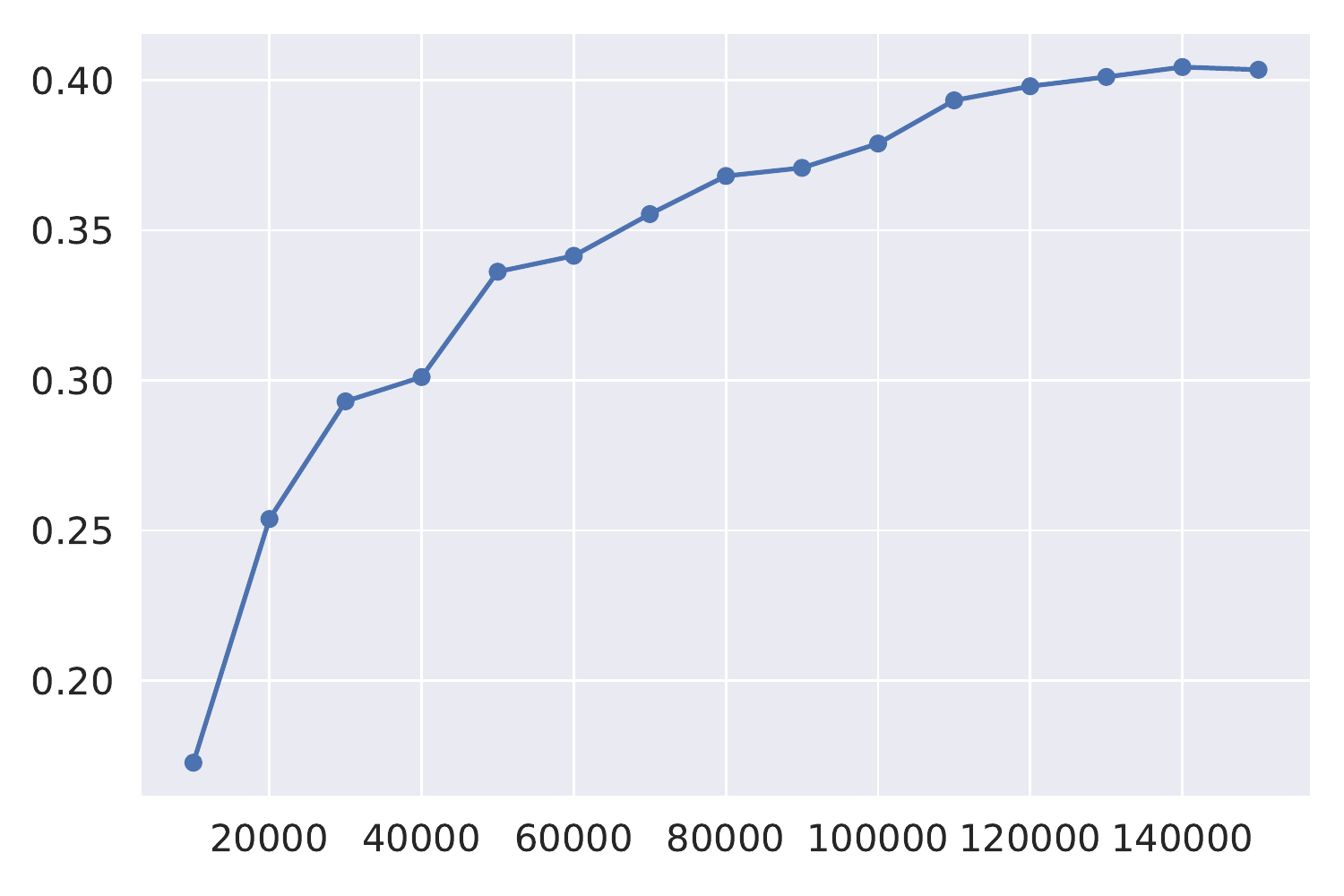}
    \caption{COCO (mAP)}
    \end{subfigure}
    \hspace{0.08\linewidth}
    \begin{subfigure}[b]{0.3\linewidth}
    \includegraphics[width=\linewidth]{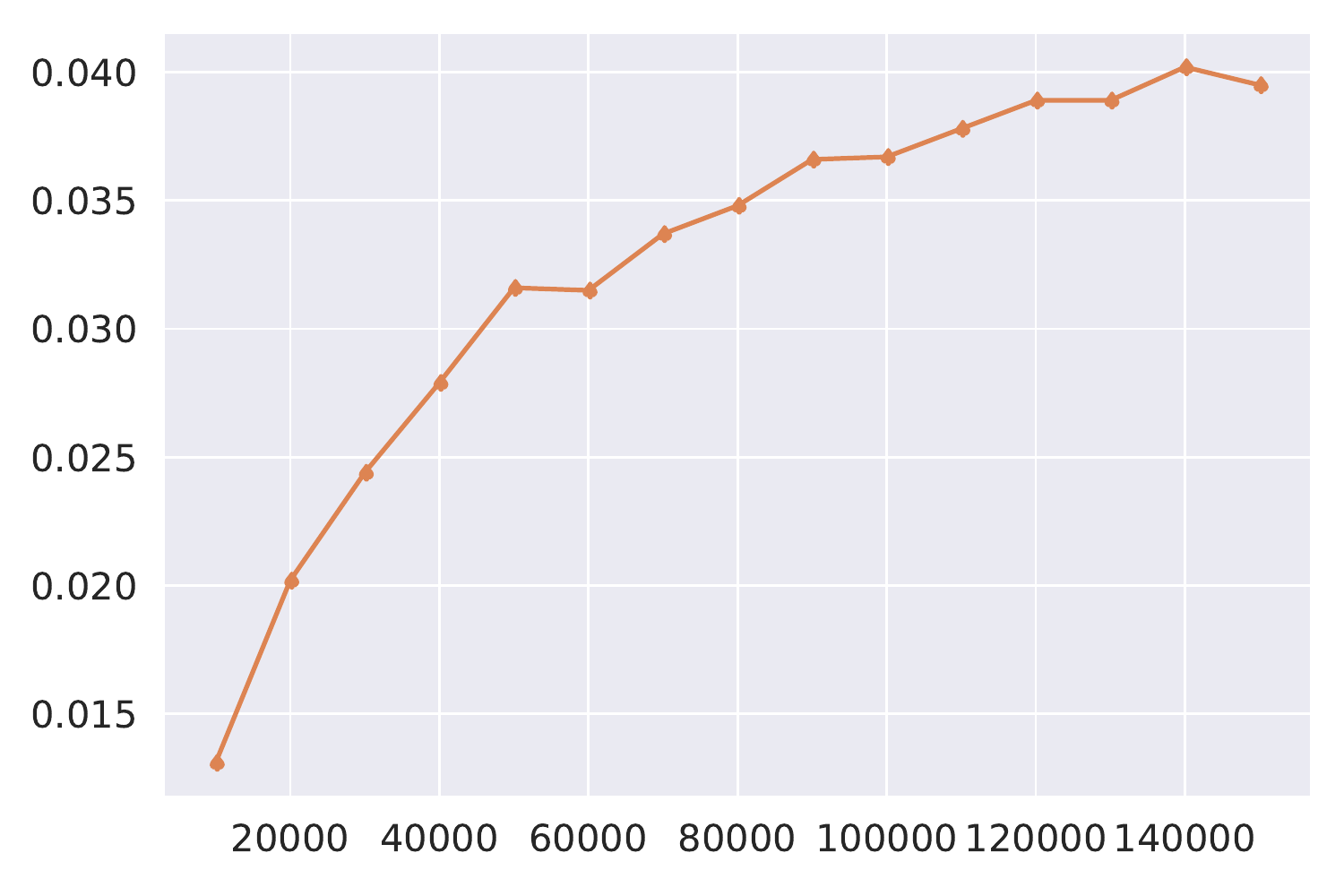}
    \caption{Visual Genome (mAP)}
    \end{subfigure}

    \begin{subfigure}[b]{0.3\linewidth}
    \includegraphics[width=\linewidth]{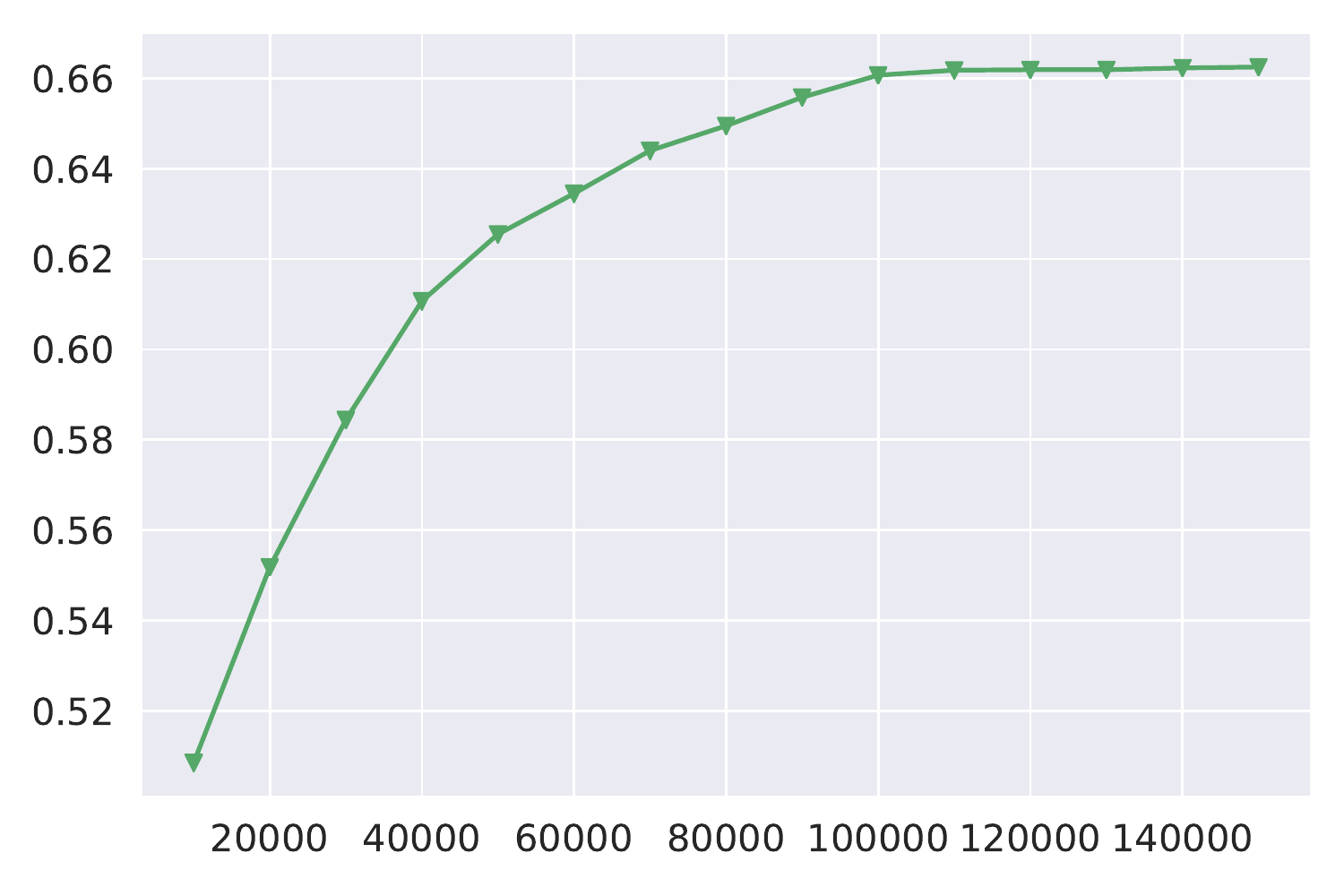}
    \caption{VQA 2.0 (accuracy)}
    \end{subfigure}
    \hspace{0.08\linewidth}
    \begin{subfigure}[b]{0.3\linewidth}
    \includegraphics[width=\linewidth]{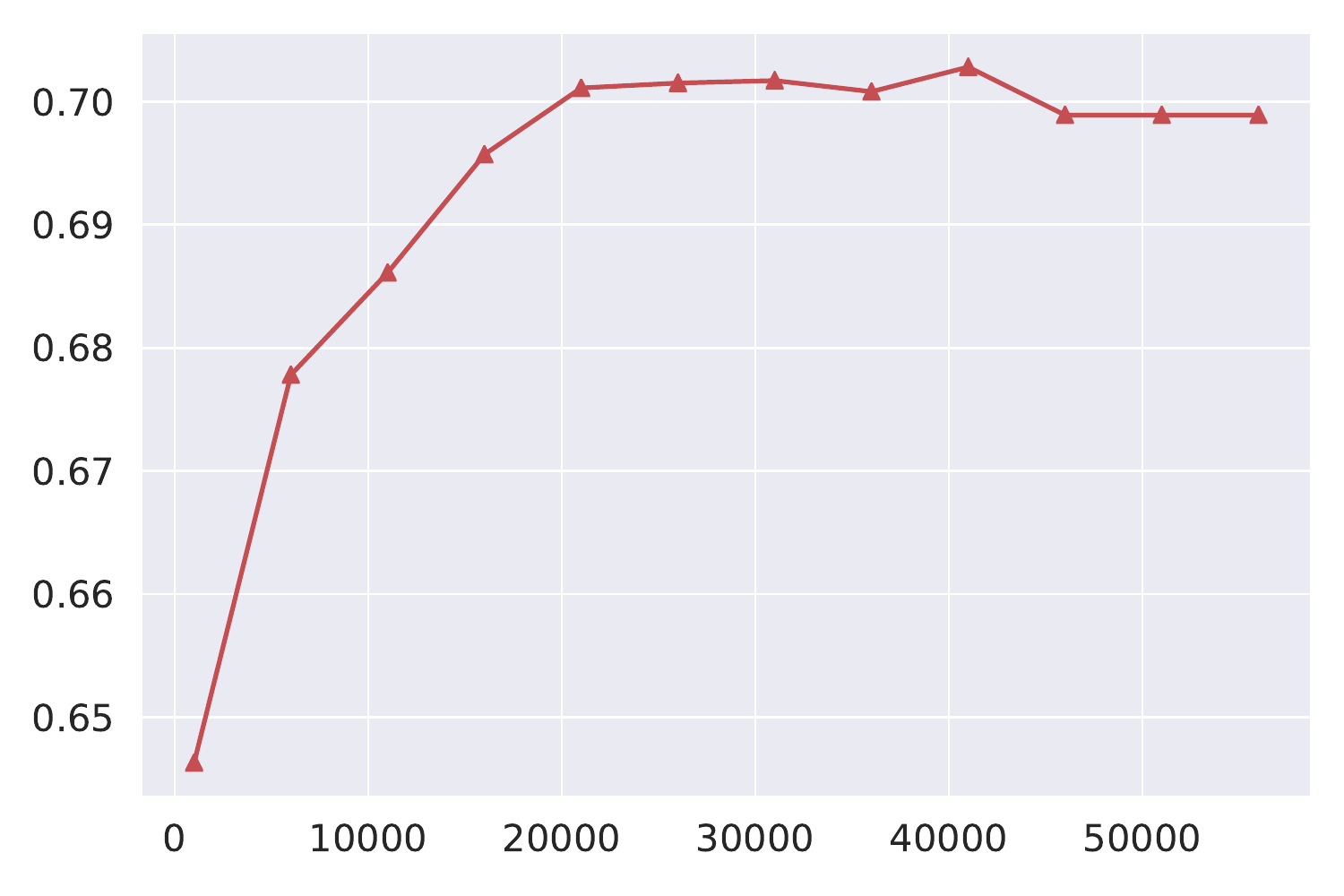}
    \caption{SNLI-VE (accuracy)}
    \end{subfigure}

    \begin{subfigure}[b]{0.3\linewidth}
    \includegraphics[width=\linewidth]{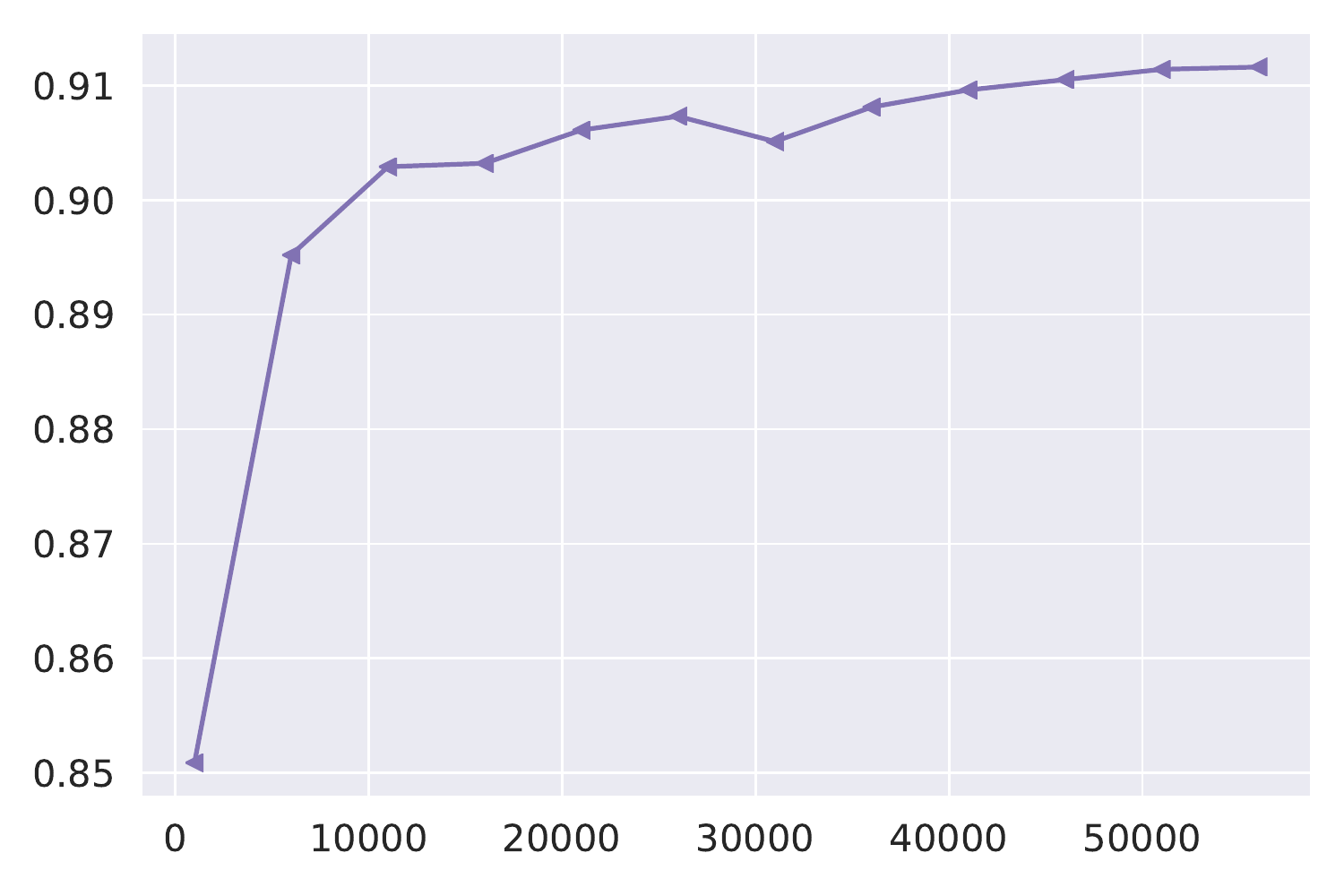}
    \caption{QQP (accuracy)}
    \end{subfigure}
    \hspace{0.08\linewidth}
    \begin{subfigure}[b]{0.3\linewidth}
    \includegraphics[width=\linewidth]{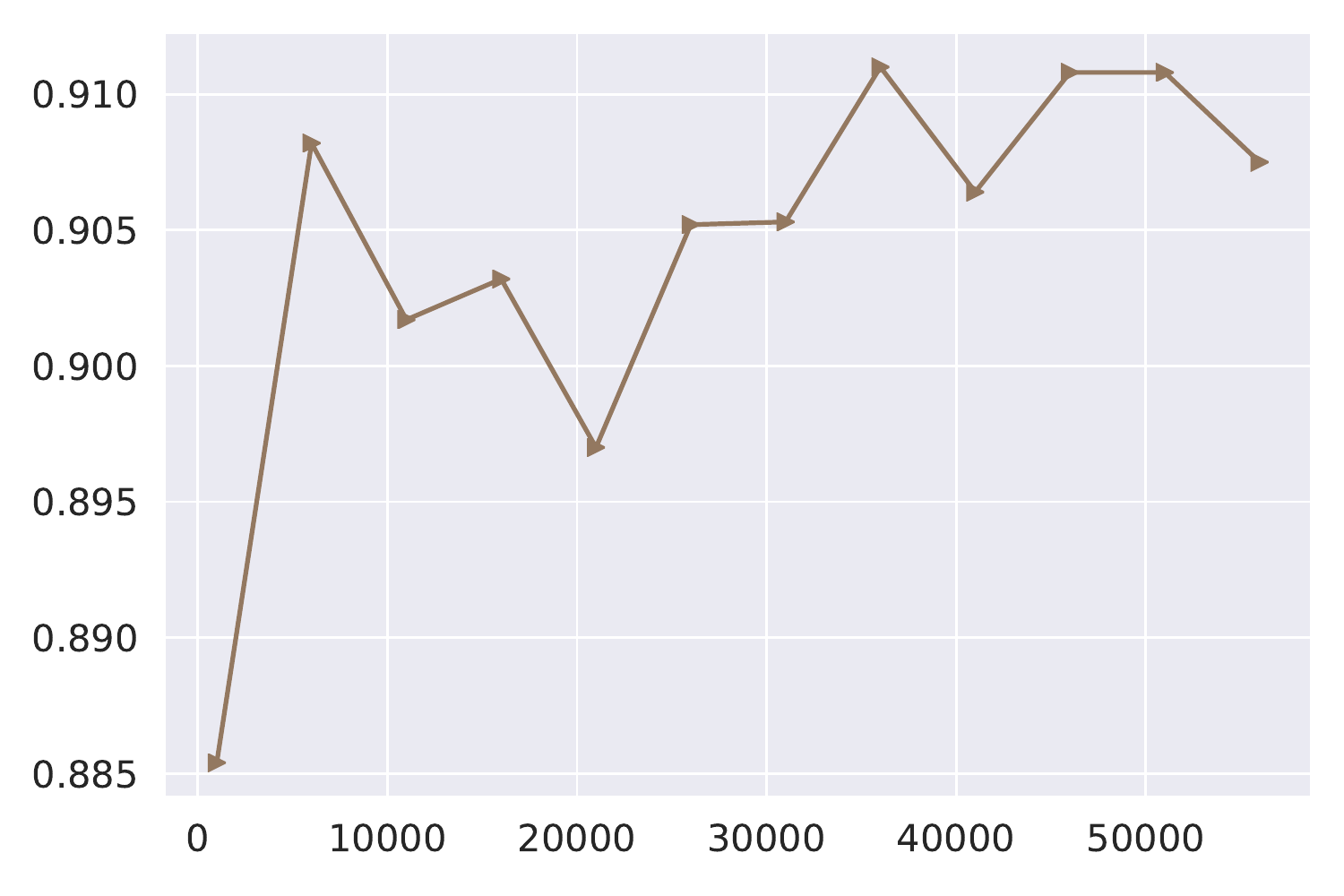}
    \caption{QNLI (accuracy)}
    \end{subfigure}

    \begin{subfigure}[b]{0.3\linewidth}
    \includegraphics[width=\linewidth]{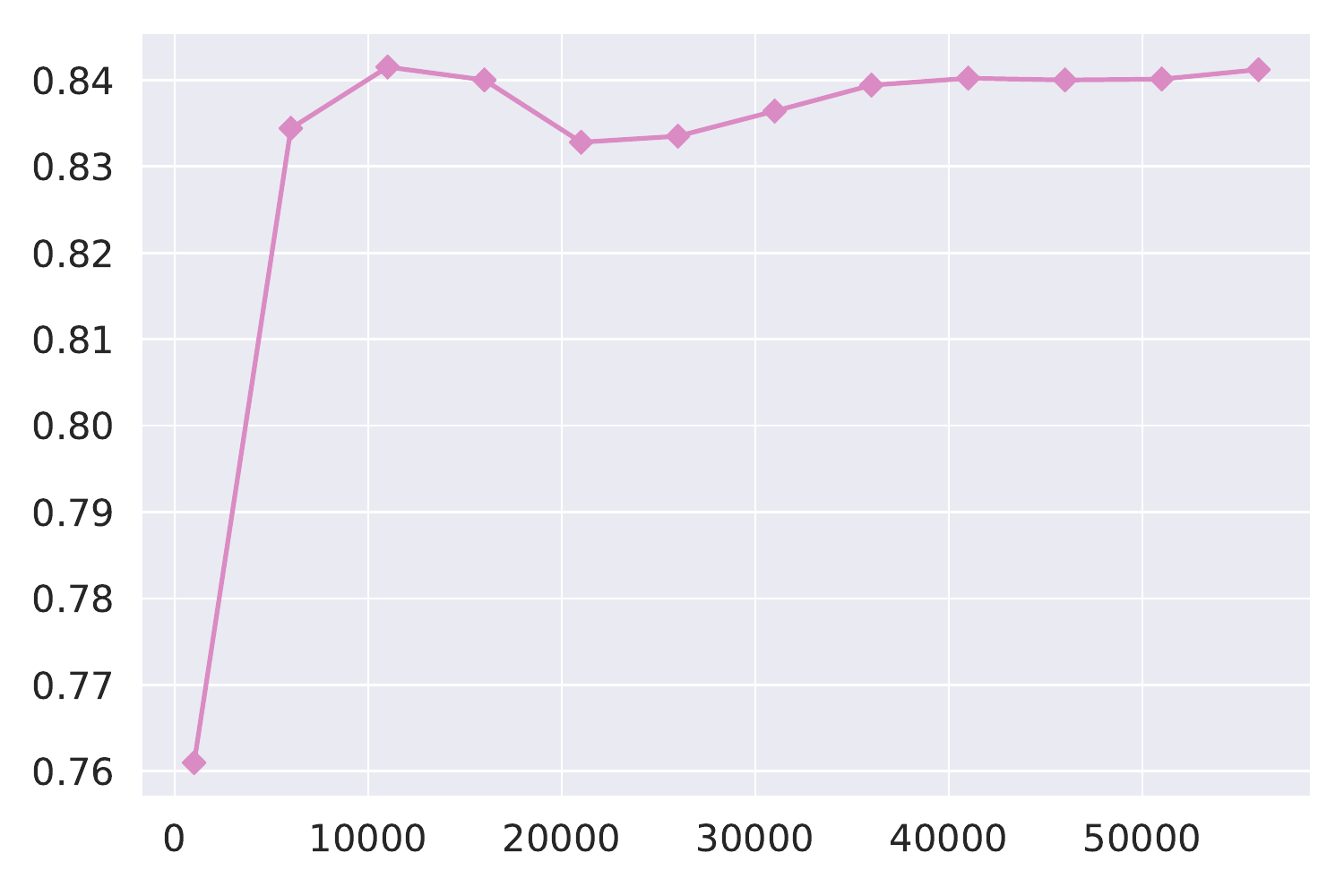}
    \caption{MNLI-mm (accuracy)}
    \end{subfigure}
    \hspace{0.08\linewidth}
    \begin{subfigure}[b]{0.3\linewidth}
    \includegraphics[width=\linewidth]{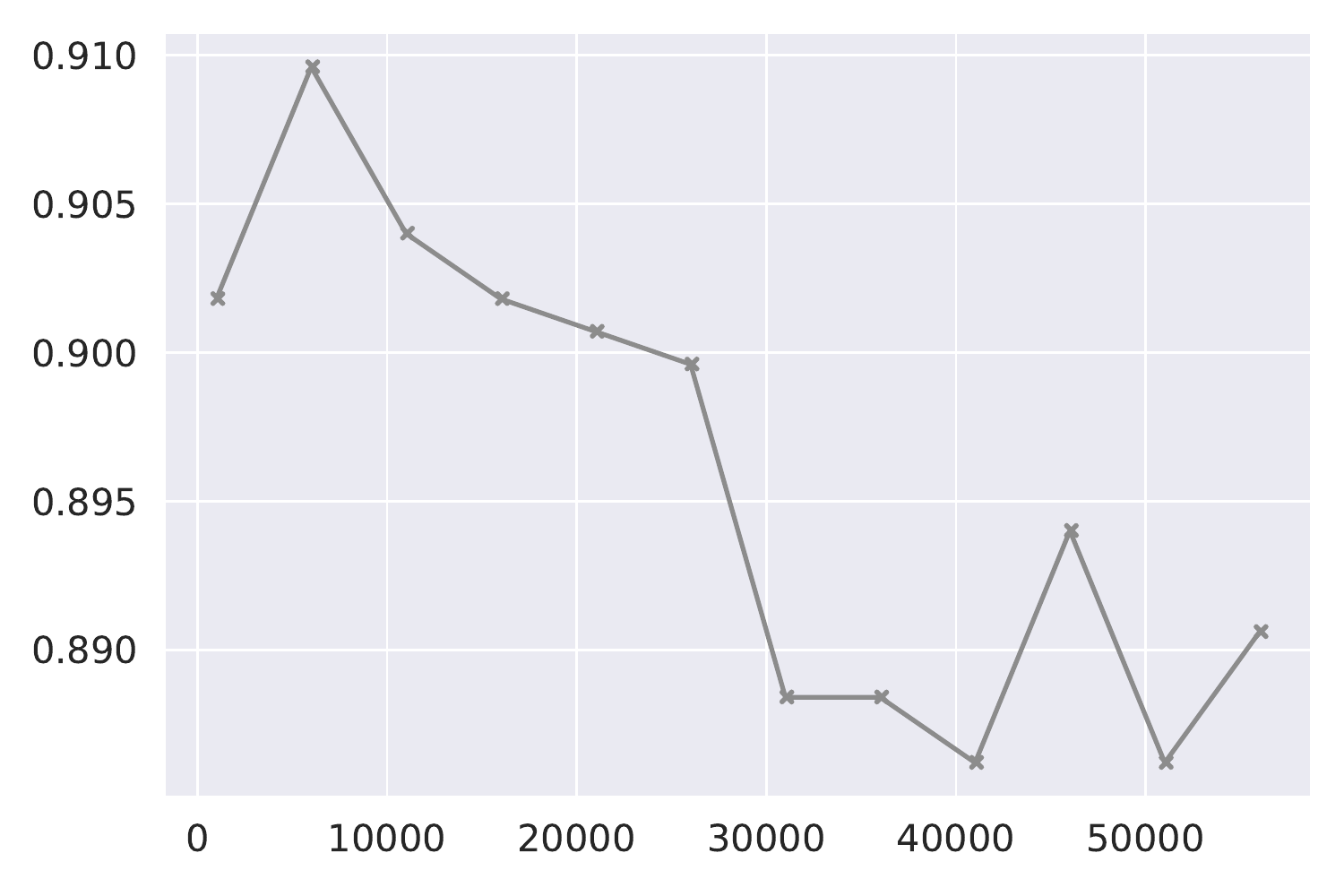}
    \caption{SST-2 (accuracy)}
    \end{subfigure}

    \caption{Learning curves of various experiments. The plots show the validation metrics at various iterations during the training process of (a) shared decoders, (b) separate decoders, and (c - j) single task training for each of the tasks.} 
    \label{fig:supp_learning_curves}
\end{figure*}

In Figure~\ref{fig:supp_learning_curves}, we show the learning curves of our unified model on all the 8 datasets with shared or separate decoders (Table~3 line 5 and 4 in the main paper), plotting the per-task performance on the validation data against training iterations. We also show the learning curves of the models trained on a single dataset (Table~3 line 1) for reference.

It can be seen that in our multi-task models, the performance of most tasks increases monotonically during training. However, SST-2 accuracy and QNLI accuracy reach their peak in early iterations and slightly decline as the training goes on, likely due to overfitting on these two relatively small datasets.

\section{More visualizations}

\begin{figure*}[t]
\vspace{-0.5em}
\centering
\includegraphics[width=\linewidth]{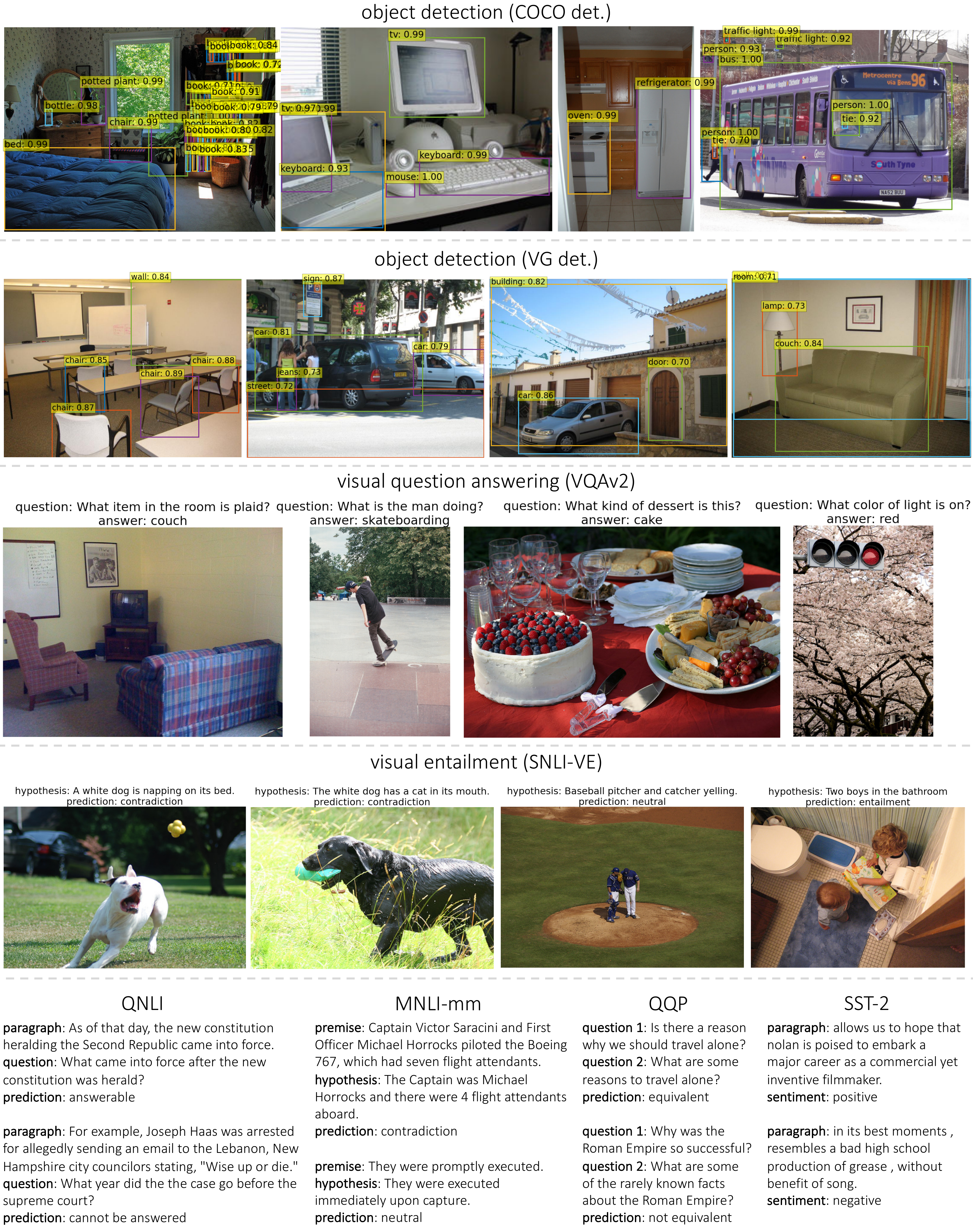}
\caption{More predictions of our model with a shared decoder (Table~3 line 5 in the main paper) across 8 datasets.}
\label{fig:supp_visualization}
\end{figure*}

Figure \ref{fig:supp_visualization} shows additional predicted examples from our \modelName model across 8 datasets (Table~3 line 5 in the main paper). The same model is applied to each task and dataset.

\end{document}